\documentclass[mathfont=cm, accepted]{uai2021} 

\usepackage[american]{babel}

\usepackage{natbib} 
\usepackage{mathtools} 
\usepackage{booktabs} 
\usepackage{tikz} 
\usepackage{amssymb}
\usepackage{bm}

\newcommand*\samethanks[1][\value{footnote}]{\footnotemark[#1]}

\title{Trusted-Maximizers Entropy Search for Efficient Bayesian Optimization}

\author[1]{
{Quoc~Phong~Nguyen\thanks{Equal contribution}}{}}
\author[3,4]{Zhaoxuan~Wu\samethanks}
\author[1]{Bryan~Kian~Hsiang~Low}
\author[2]{Patrick~Jaillet}
\affil[1]{%
    Department of Computer Science, National University of Singapore, Republic of Singapore
}
\affil[2]{%
    Department of Electrical Engineering and Computer Science, Massachusetts Institute of Technology, USA
}
\affil[3]{%
    Institute of Data Science, National University of Singapore, Republic of Singapore
}
\affil[4]{%
    NUSGS Integrative Sciences and Engineering Programme, National University of Singapore, Republic of Singapore
}

\begin{document}
\maketitle

\newcommand{\mbf}[1]{\mathbf{#1}}
\newcommand{\mcl}[1]{\mathcal{#1}}
\newcommand{\mbb}[1]{\mathbb{#1}}
\newcommand{\mds}[1]{\mathds{#1}}
\newcommand{\argmax}{\arg\!\max}

\begin{abstract}
    Information-based \emph{Bayesian optimization} (BO) algorithms have achieved state-of-the-art performance in optimizing a black-box objective function. However, they usually require several approximations or simplifying assumptions (without clearly understanding their effects on the BO performance) and/or their generalization to batch BO is computationally unwieldy, especially with an increasing batch size. To alleviate these issues, this paper presents a novel \emph{trusted-maximizers entropy search} (TES) acquisition function: It measures how much an input query contributes to the information gain on the maximizer over a finite set of \emph{trusted maximizers}, i.e., inputs optimizing functions that are sampled from the Gaussian process posterior belief of the objective function. Evaluating TES requires either only a stochastic approximation with sampling or a deterministic approximation with expectation propagation, both of which are investigated and empirically evaluated using synthetic benchmark objective functions and real-world optimization problems, e.g., hyperparameter tuning of a convolutional neural network and synthesizing `physically realizable' faces to fool a black-box face recognition system. Though TES can naturally be generalized to a batch variant with either approximation, the latter is amenable to be scaled to a much larger batch size in our experiments.
\end{abstract}

\section{Introduction}\label{sec:intro}

\emph{Bayesian optimization} (BO) is an effective strategy for iteratively obtaining noisy outputs at input queries of an \emph{objective function} to efficiently learn about the input locations of its \emph{maximum} value, i.e., its \emph{maximizers}, regardless of the function's closed-form expression, derivatives, or convexity \citep{brochu10tut,shahriari15}.
It is particularly useful when the budget of function evaluations is limited due to time/monetary costs such as in the design \citep{brochu10animate} and machine learning model training \citep{snoek12,dai2019bayesian}.
There has been a rapid growth in the number of works generalizing BO to nonmyopic BO \citep{ling16}, high-dimensional BO \citep{NghiaAAAI18}, private outsourced BO \citep{kharkovskii2020private}, multi-fidelity BO \citep{zhang2017information,zhang2019bayesian}, recursive reasoning BO \citep{dai2020r2}, and ranking BO \citep{Nguyen21topk}.

On the other hand, restricting a BO solution to only one input query at each iteration cannot take advantage of parallel function evaluations which are becoming increasingly attainable due to, for example, the rapid growth of computing resources' affordability and availability these days. In such a scenario and the time constraint, the urge to obtain the noisy outputs at a batch of input queries, i.e., \emph{batch Bayesian optimization}, becomes prominent. 
However, this task is challenging due to the extra modifications needed to account for correlation in a batch such, as seen in several existing works \citep{ginsbourger10,desautels14,ppes,gonzalez16batch,daxberger2017distributed}.  It encourages the development of BO algorithms that can be naturally generalized to the batch scenario without making additional adjustments.

To approach this problem, this paper focuses on information-based BO acquisition functions. In particular, it is the information gain about the maximizer (or the maximum value) of the objective function through observing noisy function outputs at input queries. 
This definition is much the same in spite of the number of input queries at each BO iteration, e.g., a single query or a batch of queries.
The essence of these acquisition functions is different from \emph{upper confidence bound} (UCB) that UCB minimizes the \emph{regret} (over function outputs at evaluated inputs), while information-based acquisitions aim to infer the maximizer's location regardless of the function outputs at the evaluated inputs \citep{es}.
They also have a distinctive feature from other heuristic acquisition functions such as probability of improvement \citep{kushner1964new} and \emph{expected improvement} (EI) \citep{movckus78}. It is about capturing the knowledge/belief of the maximizer (or the maximum value) over the whole input space (a global measure), which cannot be done with local heuristics, as argued in the work of \citet{es}. 
Empirically, these information-based acquisition functions such as \emph{predictive entropy search} (PES) have shown notable success in optimizing black-box functions \citep{pes}.

In this paper, we take a different approach in constructing the \emph{trusted-maximizers entropy search} (TES) acquisition function that (1) intuitively conveys the notion of maximizer without directly evaluating the information gain of the global maximizer; and (2) is amenable to batch BO. 
The first property is made feasible by the observation: while being the global maximizer requires its function output to be not less than those evaluated at all other inputs in the function's domain, 
it is much easier to eliminate an input as the global maximizer which requires the existence of only an input whose function output is larger.
Hence, if we can construct a set of ``promising'' inputs to be the maximizer, refining this set by rejecting its inputs can effectively refine the belief of the maximizer.
The second property comes from the fact that our approximation does not depend on the input queries, hence considering a batch of input queries does not increase the approximation's complexity.
Both stochastic and deterministic approximations are examined in the paper, and their empirical performances are illustrated using synthetic benchmark objective functions and real-world optimization problems, e.g., hyperparameter tuning of a convolutional neural network and synthesizing `physically realizable' faces to fool a black-box face recognition system.



%

\section{Related Works}

Most of the existing information-based works are centered around an intuitive measure of the information gain of the maximizer of the objective function, e.g., \emph{entropy search} (ES) \citep{es} and PES \citep{pes}.
While ES selects an input query by directly maximizing the information gain on the maximizer via reducing the maximizer's entropy, it requires a series of approximations. 
On the other hand, PES exploits the symmetric property of mutual information to sidestep the approximations in ES, and yet, it requires computing a predictive belief that involves an intractable constraint.
In our view, these approaches' inherent difficulty is rooted in the modeling of the objective function's global maximizer as an input whose function output is not less than those evaluated at other inputs in the domain. 
Imposing exactly this condition is hardly feasible as it involves the whole input space.
Therefore, we would like to develop TES that does not require such a condition.

There are exceptions that focus on the information gain of the maximum value of the objective function such as \emph{output-space predictive entropy search} \citep{hoffman15opes}, \emph{max-value entropy search} (MES) \citep{wang17mes}, \emph{fast information-theoretic Bayesian optimization} \citep{ru2018fast}, and \emph{binary entropy search for maximum value prediction} \citep{Nguyen21active} which result in a simpler approximation, e.g., MES enjoys a closed-form expression. However, as admitted by MES's authors, it is not intuitive that the information of the maximum value is a good search strategy for the maximizer. 

Regarding the batch BO problem, there have been several works in the existing literature such as $q$-EI with a constant liar strategy (QEI-CL) \citep{ginsbourger10}, Gaussian process batch upper confidence bound (BUCB) \citep{desautels14}, QEI \citep{marmin15}, batch BO via local penalization (BBO-LP) \citep{gonzalez16batch}, and parallel predictive entropy search (PPES) \citep{ppes}.
However, QEI-CL, BUCB, and LP-EI optimize greedily each input one by one in the batch based on a heuristic value at input queries (QEI-CL), or purely based on the predictive variance to encourage exploration (BUCB), or based on a local penalty based on an estimation of the objective function's Lipschitz constant (LP-EI). 
On the other hand, QEI and PPES are able to optimize the batch jointly, which is also a property of the proposed TES acquisition function in this paper.
It is noted that only PPES is an acquisition function that is based on the information gain.
While it is extended from the computationally efficient PES \citep{pes}, the optimization of PPES requires a computationally expensive inner loop of \emph{expectation propagation} (EP).
To the best of our knowledge, there have not been any batch BO solution extended from MES. It is probably because a direct extension of MES to a batch variant requires computing the entropy of a truncated multivariate Gaussian random variable.
\section{Problem Setup}

We propose a BO algorithm for black-box optimization problems, where the analytical form or the derivative information of the objective function are unknown. Given a black-box function $f: \mcl{X} \rightarrow \mbb{R}$, we search for the global maximizer $\mbf{x}^* \triangleq \max_{\mbf{x} \in \mcl{X}} f(\mbf{x})$ in an iterative manner. At each iteration, the BO algorithm is required to select input queries based on the posterior belief of $f$ given all observations $\mbf{y}_{\mcl{D}} \triangleq (y_{\mbf{x}'})_{\mbf{x}' \in \mcl{D}}$ at preceding input queries $\mcl{D} \subset \mcl{X}$ where $y_{\mbf{x}'} \triangleq f(\mbf{x}') + \epsilon$ and $\epsilon \sim \mcl{N}(0, \sigma_n^2)$. 
These input queries are selected to improve the belief of $\mbf{x}^*$ as much as possible. 
Specifically, BO models the posterior belief of $f$ by a GP with the function's prior mean $\mbb{E}[f(\mbf{x})]$ and covariance $k_{\mbf{x}\mbf{x}'} \triangleq \text{cov}[f(\mbf{x}), f(\mbf{x}')]$ for all $\mbf{x}, \mbf{x}' \in \mcl{X}$. The covariance is constructed with the \emph{squared exponential} (SE) kernel
$k_{\mbf{x}\mbf{x}'} \triangleq \sigma_s^2\ \exp(-0.5(\mbf{x} - \mbf{x}')^\top{\Lambda}^{-2}(\mbf{x} - \mbf{x}'))$ where its hyperparameters include the length-scales ${\Lambda} \triangleq \mathrm{diag}[\ell_1, \ldots, \ell_d]$ and the signal variance $\sigma_s^2$. The prior mean is usually set to zero to simplify the equations. Given $\mbf{y}_{\mcl{D}}$, the predictive distributions of $f(\mbf{x})$ and $y_{\mbf{x}}$ are Gaussian distributions: $p(f(\mbf{x}) |\mbf{y}_{\mathcal{D}}) \sim \mcl{N}(\mu_{\mbf{x}}, \sigma_{\mbf{x}}^2)$ and $p(y_{\mbf{x}} |\mbf{y}_{\mathcal{D}}) \sim \mcl{N}(\mu_{\mbf{x}}, \sigma_{\mbf{x}}^2 + \sigma_n^2)$, respectively, where
\begin{equation}
\begin{array}{l}
\mu_{\mbf{x}} 
	\triangleq 
K_{\mbf{x}\mcl{D}}(K_{\mcl{D}\mcl{D}}+\sigma^2_n I)^{-1}
	\mbf{y}_{\mcl{D}}\\
\sigma_{\mbf{x}}^2 
	\triangleq k_{\mbf{x} \mbf{x}}
	- K_{\mbf{x}\mcl{D}}(K_{\mcl{D}\mcl{D}}+\sigma^2_n I)^{-1}
	K_{\mcl{D}\mbf{x}}\ ,
\end{array}
\label{eq:gppost}
\end{equation}
$K_{\mbf{x}\mcl{D}}\triangleq(k_{\mbf{x}\mbf{x}'})_{\mbf{x}'\in \mcl{D}}$, $K_{\mcl{D}\mcl{D}}\triangleq(k_{\mbf{x}'\mbf{x}''})_{\mbf{x}', \mbf{x}''\in \mcl{D}}$, $I$ is the identity matrix, and $K_{\mcl{D}\mbf{x}}\triangleq K^{\top}_{\mbf{x}\mcl{D}}$.

\section{Trusted-Maximizers Entropy Search}
\begin{figure}[t]
\hspace{-4mm}
\centering
\begin{tabular}{@{}cc@{}}	
	\includegraphics[height=0.172\textwidth]{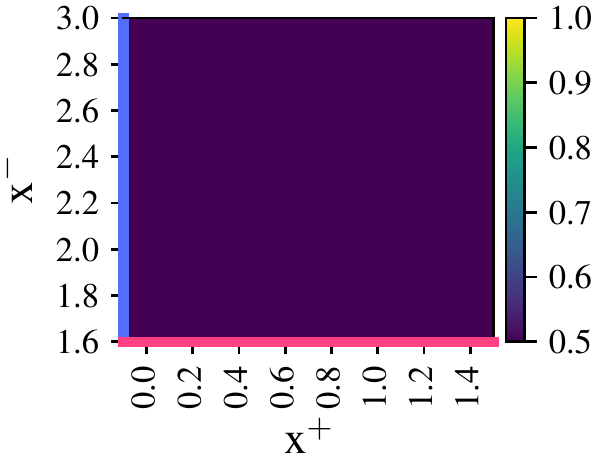}	
	&
  	\includegraphics[height=0.172\textwidth]{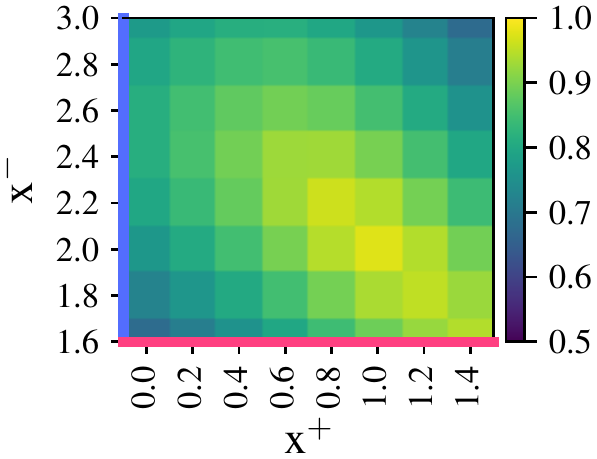}	
	\\
	(a) 
	&
	(b) 
  	\\
    \includegraphics[height=0.20\textwidth]{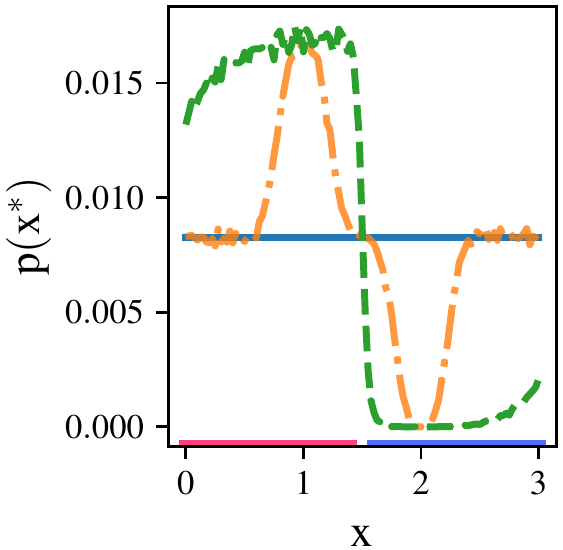}
    &
  	\includegraphics[height=0.20\textwidth]{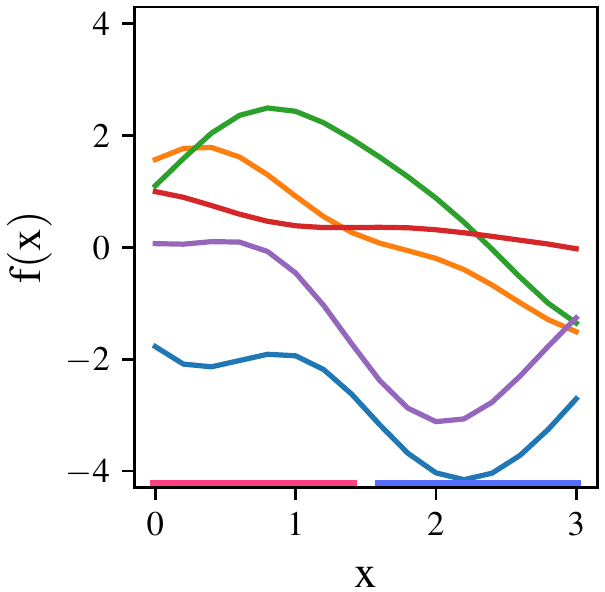}\\
  	(c) 
  	&
  	(d)
\end{tabular}
\caption{Probability of $f(\mbf{x}^+) > f(\mbf{x}^-)$
(a) before and (b) after conditioning on $f(1) > f(2)$ where $\mbf{x}^+ \in [0,1.4]$,  $\mbf{x}^- \in [1.6,3]$, and the belief of $f$ is modeled by a GP using a SE kernel with $\sigma^2_s= 5$, $\sigma^2_n= 10^{-9}$, and length-scale $\ell=1$. (c) Probability of $\mbf{x}^*$ before (blue) and after conditioning on $f(1)>f(2)$ using the same GP prior with $\ell=1$ (green) and $\ell=0.2$ (orange).
(d) $5$ functions sampled from the same GP prior with $\ell=1$ and satisfying $f(1) > f(2)$.
}
\label{fig:movex}
\end{figure}

%
%
%
%
Information-based acquisition functions like ES~\citep{es} and PES~\citep{pes} have been designed to enable their BO algorithms to improve the (posterior) belief $p(\mbf{x}^*|\mbf{y}_{\mathcal{D}})\triangleq p( f(\mbf{x}^*)~=~\max_{\mbf{x} \in \mcl{X}} f(\mbf{x})|\mbf{y}_{\mathcal{D}})$ of the global maximizer $\mbf{x}^*$ of the objective function $f$.
In particular, ES can be directly used to measure the information gain on $\mbf{x}^*$ by selecting the next input query $\mbf{x}$ for evaluating $f$ given the noisy observations $\mbf{y}_{\mcl{D}}$ from the previous BO iterations:
\begin{equation}
\alpha(\mbf{y}_{\mcl{D}},\mbf{x})
\triangleq H(\mbf{x}^*|\mbf{y}_{\mcl{D}}) - \mbb{E}_{p(y_{\mbf{x}}|\mbf{y}_{\mcl{D}})}[ H(\mbf{x}^*|\mbf{y}_{\mcl{D}\cup\{\mbf{x}\}})]\ .
\label{esacq}
\end{equation}
However, 
evaluating~\eqref{esacq} requires a series of approximations.
Furthermore, one of them incurs cubic time in the size of the discretized input domain and thus it becomes expensive with a large input domain (or risks~\eqref{esacq} being approximated poorly).
To sidestep these issues, PES exploits the symmetric property of mutual information to re-express~\eqref{esacq} as \begin{equation}
\alpha(\mbf{y}_{\mcl{D}},\mbf{x}) 
= H(y_{\mbf{x}} | \mbf{y}_{\mcl{D}}) 
- \mbb{E}_{p(\mbf{x}^*| \mbf{y}_{\mcl{D}})} [ H(y_{\mbf{x}} |\mbf{y}_{\mcl{D}}, \mbf{x}^*) ]\ .
\label{pesacq}
\end{equation} 
Intuitively, the selection of an input query $\mbf{x}$ to maximize~\eqref{pesacq} has to trade-off between exploration (hence inducing a large Gaussian predictive entropy $H(y_{\mbf{x}} | \mbf{y}_{\mcl{D}})$) vs. exploitation of the current posterior belief $p(\mbf{x}^*| \mbf{y}_{\mcl{D}})$ of  $\mbf{x}^*$ to choose a nearby input $\mbf{x}$ of $f$ to be evaluated (hence inducing a small expected predictive entropy $\mbb{E}_{p(\mbf{x}^*| \mbf{y}_{\mcl{D}})} [ H(y_{\mbf{x}} | \mbf{y}_{\mcl{D}}, \mbf{x}^*) ]$) to yield a highly informative observation that in turn improves the posterior belief of $\mbf{x}^*$.
Computing the predictive belief $p(y_{\mbf{x}}|\mbf{y}_{\mcl{D}},\mbf{x}^*)$ necessary for the second entropy term in~\eqref{pesacq} requires incorporating the constraint of $f(\mbf{x}^*) \ge f(\mbf{x})\ \forall \mbf{x} \in \mcl{X}$ which unfortunately is not tractable without imposing simplifying assumptions.
It is not clear to what extent they would compromise the quality of the resulting approximation of the predictive belief $p(y_{\mbf{x}}| \mbf{y}_{\mcl{D}}, \mbf{x}^*)$.
The work of~\citet{wang17mes} has proposed using the information gain on the global maximum $f(\mbf{x}^*)$ but admitted that it is not intuitive why it would necessarily be a good surrogate of that on the global maximizer $\mbf{x}^*$.
Therefore, this paper still focuses on finding $\mbf{x}^*$, albeit with a new perspective that is different from ES and PES, which we will discuss next.
Let $\mcl{X}^{\star}\subset\mcl{X}$ denote a \emph{finite} set of \emph{trusted maximizers}, i.e., inputs that are more likely to be the global maximizer $\mbf{x}^*$ in the current BO iteration, that is, $p(\mbf{x}^* \in \mcl{X}^{\star} | \mbf{y}_{\mcl{D}})$ is large. 
Let $\mbf{x}^{\star}$ be a r.v. representing the maximizer of $f$ over $\mcl{X}^{\star}$, that is, $p(\mbf{x}^{\star}| \mbf{y}_{\mcl{D}}) \triangleq p( f(\mbf{x}^{\star}) = \max_{\mbf{x} \in \mcl{X}^{\star}} f(\mbf{x})| \mbf{y}_{\mcl{D}})$.
To ease the exposition of our key idea here, we consider a unique maximizer of $f$ over $\mcl{X}^{\star}$ but do not assume this when we describe our technical approach later.
By finding the maximizer $\mbf{x}^{\star}$ of $f$ over $\mcl{X}^{\star}$, it effectively eliminates the remaining input queries in $\mcl{X}^{\star} \setminus \{ \mbf{x}^{\star} \}$ from being the global maximizer $\mbf{x}^*$ because $f(\mbf{x}^*) \ge f(\mbf{x}^{\star}) > f(\mbf{x})$ for all $\mbf{x} \in \mcl{X}^{\star} \setminus \{ \mbf{x}^{\star} \}$.
So, if $p(\mbf{x}^* \in \mcl{X}^{\star} | \mbf{y}_{\mcl{D}})$ is large, then eliminating some input queries in $\mcl{X}^{\star}$ would gain us useful information on the global maximizer $\mbf{x}^*$.
Furthermore, besides reducing the probability of each input query $\mbf{x} \in \mcl{X}^{\star} \setminus \{ \mbf{x}^{\star} \}$ being $\mbf{x}^*$, 
their neighboring input queries  also experience a reduced probability of being $\mbf{x}^*$.
We illustrate this with a simple example of input dimension $d=1$ in Fig.~\ref{fig:movex}:
Let $\mbf{x}^+ \in [0, 1.4]$ and $\mbf{x}^- \in [1.6, 3]$.
It can be observed that by conditioning on only $f(1) > f(2)$, 
the probability of $f(\mbf{x}^+)> f(\mbf{x}^-)$ in Fig.~\ref{fig:movex}b (hence implying $f(\mbf{x}^*) > f(\mbf{x}^-)$) increases from that of $0.5$ in Fig.~\ref{fig:movex}a.
In other words, the probability of $\mbf{x}^- \in [1.6, 3]$ being $\mbf{x}^*$ reduces. The magnitude of reduction depends on the correlation between $f(2)$ and the outputs of $f$ evaluated at the input queries in $[1.6, 3]$, which in turn depends on the length-scale of the SE kernel (see Fig.~\ref{fig:movex}c). 
%

So, if we can construct the set $\mcl{X}^{\star}$ of trusted maximizers such that $p(\mbf{x}^* \in \mcl{X}^{\star}|\mbf{y}_{\mcl{D}})$ is large, then 
finding the maximizer $\mbf{x}^{\star}$ of $f$ over $\mcl{X}^{\star}$ would
gain us useful information on the global maximizer $\mbf{x}^*$ since it reduces the probability of each input query $\mbf{x}\in\mcl{X}^{\star} \setminus \{\mbf{x}^{\star}\}$ and its neighbors being $\mbf{x}^*$, as explained above. 
We therefore form
$\mcl{X}^{\star}$    
using samples of $\mbf{x}^*$ drawn from the posterior belief $p(\mbf{x}^*|\mbf{y}_{\mcl{D}})$: 
To do this, we adopt the method of~\citet{pes} to first approximately sample $|\mcl{X}^{\star}|$ functions from the GP predictive belief given $\mbf{y}_{\mcl{D}}$~\eqref{eq:gppost} and then compute the global maximizers of these sampled functions to form $\mcl{X}^{\star}$.
%
%

Given the set $\mcl{X}^{\star}$ of trusted maximizers computed in the current BO iteration, our \emph{trusted-maximizers entropy search} (TES) acquisition function is designed to enable its BO algorithm to find the maximizer $\mbf{x}^{\star}$ of $f$ over $\mcl{X}^{\star}$ by improving the posterior belief $p(\mbf{x}^{\star}|\mbf{y}_{\mcl{D}})$.
Specifically, TES measures the information gain
on $\mbf{x}^{\star}$ by selecting the next input query $\mbf{x}\in\mcl{X}$ for evaluating $f$ given the noisy observations $(\mcl{D},\mbf{y}_{\mcl{D}})$ from the previous BO iterations:
\begin{equation}
\alpha^{\star}(\mbf{y}_{\mcl{D}},\mbf{x})
\triangleq H(\mbf{x}^{\star}|\mbf{y}_{\mcl{D}}) - \mbb{E}_{p(y_{\mbf{x}}|\mbf{y}_{\mcl{D}})}[ H(\mbf{x}^{\star}|\mbf{y}_{\mcl{D}\cup\{\mbf{x}\}})]\ .\hspace{-0.2mm}
\label{msesacq}
\end{equation}
Since the 
second entropy term in~\eqref{msesacq} requires
the posterior belief $p(\mbf{x}^{\star} | \mbf{y}_{\mcl{D}\cup\{\mbf{x}\}})$ to be computed for prohibitively many different noisy outputs $y_{\mbf{x}}$, we exploit the symmetric property of  mutual information to re-express~\eqref{msesacq} as
%
\begin{equation}
\hspace{-1.7mm}
\begin{array}{l}
\displaystyle\alpha^{\star}(\mbf{y}_{\mcl{D}},\mbf{x}) 
\hspace{-0.5mm}=\hspace{-0.5mm} H(y_{\mbf{x}} | \mbf{y}_{\mcl{D}}) \hspace{-0.2mm}- \hspace{-0.7mm}
\sum_{\mbf{x}^{\star} \in \mcl{X}^{\star}} p(\mbf{x}^{\star}| \mbf{y}_{\mcl{D}}) H(y_{\mbf{x}} |\mbf{y}_{\mcl{D}}, \mbf{x}^{\star})\vspace{0.5mm}\\
\displaystyle H(y_{\mbf{x}} |\mbf{y}_{\mcl{D}}, \mbf{x}^{\star}) \triangleq \hspace{-0.5mm}-\hspace{-1mm} \int p(y_{\mbf{x}}| \mbf{y}_{\mcl{D}}, \mbf{x}^{\star}) \log p(y_{\mbf{x}}| \mbf{y}_{\mcl{D}}, \mbf{x}^{\star})\ \text{d}y_{\mbf{x}}
\end{array}
\label{eq:soc}
\end{equation}
where $p(\mbf{x}^{\star}|\mbf{y}_{\mcl{D}})$ can be evaluated efficiently (Appendix~\ref{app:opmf}).
\paragraph{\textbf{Remark 1.}} TES~\eqref{eq:soc} differs from PES~\eqref{pesacq} in measuring the information gain on the maximizer $\mbf{x}^{\star}$ of $f$ over a \emph{finite} $\mcl{X}^{\star}$ rather than the entire infinite input domain $\mcl{X}$, which we have motivated earlier in this section. A practical implication is that PES needs to assume simplified constraints to achieve tractability while TES does not.
For example, PES draws approximate posterior samples of $\mbf{x}^*$
but does not directly exploit the constraint that the global maximum is not less than the outputs of $f$ evaluated at these samples. Instead, it assumes a simplified soft constraint (with Gaussian noise $\epsilon$) that the output of $f$ evaluated at each sample is not less than the largest noisy output observed in previous BO iterations.
%
By observing that $p(y_{\mbf{x}}| \mbf{f}_{\mcl{X}^{\star}}, \mbf{y}_{\mcl{D}}, \mbf{x}^{\star}) = p(y_{\mbf{x}}| \mbf{f}_{\mcl{X}^{\star}}, \mbf{y}_{\mcl{D}})$, the predictive belief $p(y_{\mbf{x}}| \mbf{y}_{\mcl{D}},\mbf{x}^{\star})$ in the $H(y_{\mbf{x}} |\mbf{y}_{\mcl{D}}, \mbf{x}^{\star})$ term~\eqref{eq:soc} can be derived by marginalizing out $\mbf{f}_{\mcl{X}^{\star}}\triangleq (f(\mbf{x}'))^{\top}_{\mbf{x}' \in \mcl{X}^{\star}}$:
\begin{equation}
\hspace{-0mm}
p(y_{\mbf{x}}| \mbf{y}_{\mcl{D}},\mbf{x}^{\star})
	= \int p(y_{\mbf{x}}| \mbf{f}_{\mcl{X}^{\star}}, \mbf{y}_{\mcl{D}})
	  \ p(\mbf{f}_{\mcl{X}^{\star}}| \mbf{y}_{\mcl{D}}, \mbf{x}^{\star})\ \text{d}\mbf{f}_{\mcl{X}^{\star}} .\hspace{-1.6mm}
\label{eq:yposto}
\end{equation}
Since $p(\mbf{f}_{\mcl{X}^{\star}}| \mbf{y}_{\mcl{D}},\mbf{x}^{\star})$ cannot be evaluated in closed form, we approximate it with either (a) sampling whose approximation quality improves with the number of samples or (b) a Gaussian distribution to make $p(y_{\mbf{x}}| \mbf{y}_{\mcl{D}},\mbf{x}^{\star})$ Gaussian. These two approximation methods are discussed next. 

\paragraph{\textbf{Remark 2.}} Since $p(\mbf{f}_{\mcl{X}^{\star}}| \mbf{y}_{\mcl{D}}, \mbf{x}^{\star})$  
does not depend on the input query $\mbf{x}$, its approximation remains unchanged when our TES acquisition function~\eqref{eq:soc} is optimized
with respect to the input query $\mbf{x}\in\mcl{X}$ in each BO iteration, that is, $\max_{\mbf{x}\in\mcl{X}}\alpha^{\star}(\mbf{y}_{\mcl{D}}, \mbf{x})$. So, $p(\mbf{f}_{\mcl{X}^{\star}}| \mbf{y}_{\mcl{D}}, \mbf{x}^{\star})$ is approximated only once 
in each BO iteration, which simplifies the optimization procedure and facilitates an efficient generalization to batch BO (Section~\ref{bbo}). In contrast, the approximations needed to derive PPES and ES depend on $\mbf{x}$ and hence have to be repeated for every $\mbf{x}\in\mcl{X}$ to optimize PPES and ES.

We will discuss  the evaluation of TES by both stochastic approximation with sampling, denoted as TES\textsubscript{sp}, and deterministic approximation with \emph{expectation propagation} (EP), denoted as TES\textsubscript{ep}, as below.

\subsection{Stochastic Approximation with Sampling}
\label{subsec:sampling}
Let $\mcl{F}_{\mbf{x}^{\star}}$ denote a set of samples that are randomly drawn from $p(\mbf{f}_{\mcl{X}^{\star}}|\mbf{y}_{\mcl{D}}, \mbf{x}^{\star})$. This can be obtained via rejection sampling by drawing samples from the GP predictive belief $p(\mbf{f}_{\mcl{X}^{\star}}|\mbf{y}_{\mcl{D}})$~\eqref{eq:gppost} and rejecting those samples of $\mbf{f}_{\mcl{X}^{\star}}$ where 
$\max_{\mbf{x}' \in \mcl{X}^{\star}} f(\mbf{x}')> f(\mbf{x}^{\star})$.
However, if $p(\mbf{x}^{\star}|\mbf{y}_{\mcl{D}})$ is small, then such a sampling becomes inefficient due to a large number of rejected samples. 
Importance sampling can be used to avoid this issue. 
Let $\mbf{f}_{\setminus \star}\triangleq (f(\mbf{x}'))^{\top}_{\mbf{x}' \in \mcl{X}^{\star}\setminus\{\mbf{x}^{\star}\}}$. Then, 
$$
\begin{array}{l}
p(\mbf{f}_{\mcl{X}^{\star}}| \mbf{y}_{\mcl{D}}, \mbf{x}^{\star}) \propto p(\mbf{f}_{\mcl{X}^{\star}}, \mbf{x}^{\star} | \mbf{y}_{\mcl{D}}) \vspace{0.5mm}\\
=p(\mbf{x}^{\star} | \mbf{f}_{\setminus \star}, \mbf{y}_{\mcl{D}})\ 
p(\mbf{f}_{\setminus \star} | \mbf{y}_{\mcl{D}})\ 
p(f(\mbf{x}^{\star}) | \mbf{f}_{\setminus \star}, \mbf{y}_{\mcl{D}}, \mbf{x}^{\star})\ .
\end{array}
$$
We first draw samples of $\mbf{f}_{\mcl{X}^{\star}}$ from $p(\mbf{f}_{\setminus \star} | \mbf{y}_{\mcl{D}})$ $p(f(\mbf{x}^{\star}) | \mbf{f}_{\setminus \star}, \mbf{y}_{\mcl{D}},\mbf{x}^{\star})$ and then weight these samples by 
$p(\mbf{x}^{\star} | \mbf{f}_{\setminus \star}, \mbf{y}_{\mcl{D}})  = 1 - \Phi_{ p(f(\mbf{x}^{\star})| \mbf{f}_{\setminus \star}, \mbf{y}_{\mcl{D}})} ( f^+ )$ where $f^+\triangleq\max_{\mbf{x}' \in \mcl{X}^{\star}\setminus\{\mbf{x}^{\star}\}} f(\mbf{x}')$ and $\Phi_{ p(f(\mbf{x}^{\star})| \mbf{f}_{\setminus \star}, \mbf{y}_{\mcl{D}})} ( f^+)$ is the \emph{cumulative distribution function} (c.d.f.) of the GP predictive belief $p(f(\mbf{x}^{\star})| \mbf{f}_{\setminus \star}, \mbf{y}_{\mcl{D}})$ evaluated at $f^+$. The details of importance sampling are given in Appendix~\ref{app:imppostcand}. 

Alternatively, if the number of samples to be drawn from $p(\mbf{f}_{\mcl{X}^{\star}}| \mbf{y}_{\mcl{D}}, \mbf{x}^{\star})$ is allowed to vary for different $\mbf{x}^{\star}$, then we can simply draw samples from the GP predictive belief $p(\mbf{f}_{\mcl{X}^{\star}}|\mbf{y}_{\mcl{D}})$ and group them based on their maximizers. This may result in no sample for some $\mbf{x}^{\star}$, especially if $p(\mbf{x}^{\star}| \mbf{y}_{\mcl{D}})$ is small. In this case, we simply update $\mcl{X}^{\star}$ to omit such $\mbf{x}^{\star}$. 
It follows from Remark $2$ that the construction of $\mcl{F}_{\mbf{x}^{\star}}$ does not depend on the input query $\mbf{x}$ and is hence performed only once in each BO iteration.

Given $\mcl{F}_{\mbf{x}^{\star}}$, $p(y_{\mbf{x}}|\mbf{y}_{\mcl{D}}, \mbf{x}^{\star})$ in~\eqref{eq:yposto} can be approximated by
%
$$
\begin{array}{c}
q_{\text{sp}}(y_{\mbf{x}}|\mbf{y}_{\mcl{D}}, \mbf{x}^{\star}) \triangleq |\mcl{F}_{\mbf{x}^{\star}}|^{-1} \sum_{\mbf{f}_{\mcl{X}^{\star}} \in \mcl{F}_{\mbf{x}^{\star}}} p(y_{\mbf{x}} | \mbf{f}_{\mcl{X}^{\star}}, \mbf{y}_{\mcl{D}}) 
\end{array}
$$
%
where the GP predictive belief $p(y_{\mbf{x}} | \mbf{f}_{\mcl{X}^{\star}}, \mbf{y}_{\mcl{D}})$ is derived in Appendix~\ref{app:postygvsam}. Then, TES\textsubscript{sp} can be evaluated as follows
\begin{align*}
&\alpha_{\text{sp}}^\star(\mbf{y}_{\mcl{D}}, \mbf{x}) \triangleq |\mcl{F}_{\mbf{x}^\star}|^{-1} \sum_{\mbf{x}^\star \in \mcl{X}^\star} p(\mbf{x}^\star|\mbf{y}_{\mcl{D}}) \sum_{\mbf{f}_{\mcl{X}^{\star}} \in \mcl{F}_{\mbf{x}^{\star}}}\nonumber\\ 
&\quad \mbb{E}_{p(y_{\mbf{x}} | \mbf{f}_{\mcl{X}^{\star}}, \mbf{y}_{\mcl{D}})} (
\log q_{\text{sp}}(y_{\mbf{x}}|\mbf{y}_{\mcl{D}}, \mbf{x}^{\star})
- \log q_{\text{sp}}(y_{\mbf{x}}| \mbf{y}_{\mcl{D}})
)
\end{align*}
%
where $q_{\text{sp}}(y_{\mbf{x}}| \mbf{y}_{\mcl{D}}) \triangleq \sum_{\mbf{x}^{\star} \in \mcl{X}^{\star}} p(\mbf{x}^{\star}|\mbf{y}_{\mcl{D}})  q_{\text{sp}}(y_{\mbf{x}}|\mbf{y}_{\mcl{D}}, \mbf{x}^{\star})$.
%
%
%
It is amenable to stochastic optimization 
with respect to $\mbf{x}$ via stochastic gradients which are derived using mini-batches of $y_{\mbf{x}}$ randomly sampled from the GP predictive belief $p(y_{\mbf{x}} | \mbf{f}_{\mcl{X}^{\star}}, \mbf{y}_{\mcl{D}})$ for all $\mbf{f}_{\mcl{X}^{\star}} \in \mcl{F}_{\mbf{x}^{\star}}$\footnote{Reparameterization trick is used \citep{kingma2013auto}}.

Given $\mcl{X}^\star$, the time complexity to evaluate $\alpha_{\text{sp}}^\star(\mbf{y}_{\mcl{D}},\mbf{x})$ is $\mcl{O}(|\mcl{D}|^2 |\mcl{X}^\star| + |\mcl{F}_{\mbf{x}^\star}| |\mcl{X}^\star|^2 + |\mcl{X}^\star|^3 + (|\mcl{X}^\star| + |\mcl{D}|)^3)$ where $\mcl{O}(|\mcl{D}|^2 |\mcl{X}^\star|)$ is the time complexity of the GP prediction at $\mcl{X}^\star$ given $\mbf{y}_{\mcl{D}}$, $\mcl{O}(|\mcl{F}_{\mbf{x}^\star}| |\mcl{X}^\star|^2 + |\mcl{X}^\star|^3)$ is the time complexity of sampling $\mcl{F}_{\mbf{x}^\star}$, and $\mcl{O}((|\mcl{X}^\star| + |\mcl{D}|)^3)$ is the time complexity of evaluating $p(\mbf{y}_{\mbf{x}}|\mbf{f}_{\mcl{X}^\star}, \mbf{y}_{\mcl{D}})$.


%

\subsection{Deterministic Approximation with Expectation Propagation (EP)}
\begin{figure}[t]
\begin{tabular}{@{}ccc@{}}
	\includegraphics[height=0.125\textwidth]{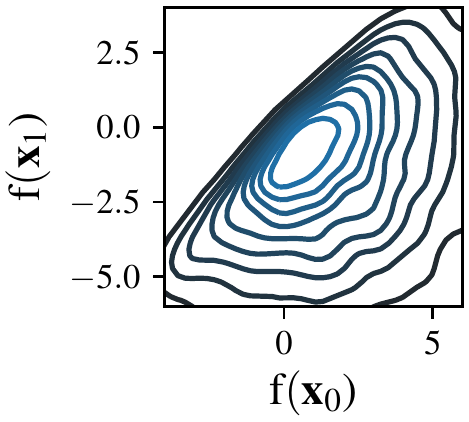}\hspace{-1mm}
	&
	\includegraphics[height=0.125\textwidth]{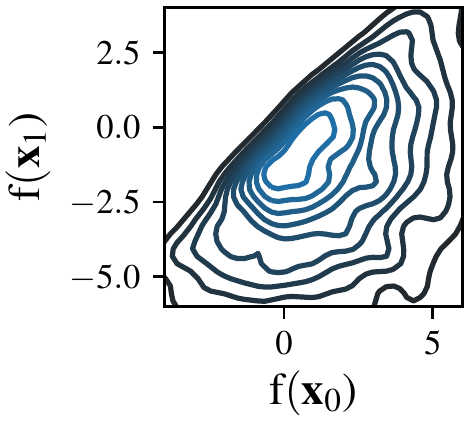}\hspace{-1mm}
	&
	\includegraphics[height=0.125\textwidth]{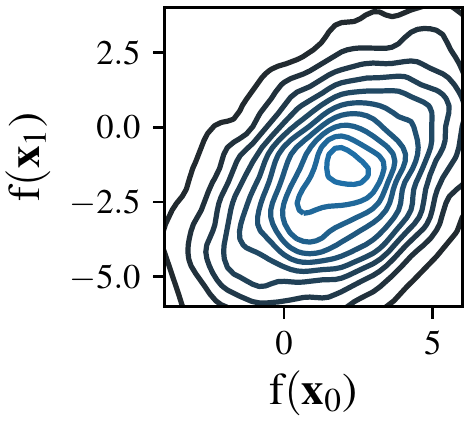}
	\\
	(a)\hspace{-1mm}
	&
	(b)\hspace{-1mm}
	&
	(c)
\end{tabular}
\caption{$p(f(\mbf{x}_0), f(\mbf{x}_1) | f(\mbf{x}_0) > f(\mbf{x}_1))$ produced by (a) rejection sampling, (b) importance sampling, and (c) EP where $p(f(\mbf{x}_0), f(\mbf{x}_1))$ is a standard bivariate Gaussian.}
\label{fig:approxdist}
\end{figure}
Alternatively, we can approximate $p(\mbf{f}_{\mcl{X}^{\star}} | \mbf{y}_{\mcl{D}}, \mbf{x}^{\star})$ with a Gaussian distribution so that $p(y_{\mbf{x}}| \mbf{y}_{\mcl{D}},\mbf{x}^{\star})$~\eqref{eq:yposto} can be evaluated in closed form.
Compared with the above sampling method, its approximation cannot be refined due to its restriction to a Gaussian~(see Fig.~\ref{fig:approxdist}c).
Nevertheless, it is time- and memory-efficient.
Its approximation is performed by matching the moments: The mean and covariance of the  Gaussian distribution are the same as that of $p(\mbf{f}_{\mcl{X}^{\star}} | \mbf{y}_{\mcl{D}}, \mbf{x}^{\star})$ being approximated. 
To do this, one can compute the empirical mean and covariance of the samples in $\mcl{F}_{\mbf{x}^{\star}}$. An alternative without the sampling of $p(\mbf{f}_{\mcl{X}^{\star}} | \mbf{y}_{\mcl{D}}, \mbf{x}^{\star})$ is EP~\citep{minka01ep} which approximates $p(\mbf{f}_{\mcl{X}^{\star}} | \mbf{y}_{\mcl{D}}, \mbf{x}^{\star})$ with a Gaussian $q_{\text{ep}}(\mbf{f}_{\mcl{X}^{\star}}|\mbf{y}_{\mcl{D}},\mbf{x}^{\star}) \triangleq \mathcal{N}(\bm{\mu}_{\text{ep}}, \bm{\Sigma}_{\text{ep}})$, as detailed in Appendix~\ref{app:eppostcand}. 
It follows from Remark $2$ that 
$q_{\text{ep}}(\mbf{f}_{\mcl{X}^{\star}}|\mbf{y}_{\mcl{D}},\mbf{x}^{\star})$ does not depend on the input query $\mbf{x}$ and is hence computed only once in each BO iteration.
Consequently, $p(y_{\mbf{x}}| \mbf{y}_{\mcl{D}},\mbf{x}^{\star})$~\eqref{eq:yposto} can be approximated in closed form with a Gaussian:
\begin{equation}
q_{\text{ep}}(y_{\mbf{x}}|\mbf{y}_{\mcl{D}},\mbf{x}^{\star}) \triangleq\mathcal{N}(\mbf{a}^\top \bm{\mu}_{\text{ep}} + b, \sigma^2_{\mbf{x}|\mcl{X}^{\star}} + \mbf{a}^\top \bm{\Sigma}_{\text{ep}} \mbf{a} + \sigma_n^2)
\label{eq:epapprox}
\end{equation}
where the posterior variance $\sigma^2_{\mbf{x}|\mcl{X}^{\star}}$ of the GP predictive belief $p(f(\mbf{x})|\mbf{f}_{\mcl{X}^{\star}}, \mbf{y}_{\mcl{D}})$, $\mbf{a}$, and $b$ are given in Appendix~\ref{app:eppostpred}. 
Then, TES\textsubscript{ep} can be evaluated as follows
\begin{align*}
&\alpha_{\text{ep}}^\star(\mbf{y}_{\mcl{D}}, \mbf{x}) \triangleq  \sum_{\mbf{x}^\star \in \mcl{X}^\star} p(\mbf{x}^\star|\mbf{y}_{\mcl{D}}) \\
&\quad\mbb{E}_{q_{\text{ep}}(y_{\mbf{x}}| \mbf{y}_{\mcl{D}}, \mbf{x}^\star)} \left(
\log q_{\text{ep}}(y_{\mbf{x}}| \mbf{y}_{\mcl{D}}, \mbf{x}^\star)
-
\log q_{\text{ep}}(y_{\mbf{x}}| \mbf{y}_{\mcl{D}})
\right)
\end{align*}
%
where $q_{\text{ep}}(y_{\mbf{x}}| \mbf{y}_{\mcl{D}}) \triangleq \sum_{\mbf{x}^{\star} \in \mcl{X}^{\star}} p(\mbf{x}^{\star}| \mbf{y}_{\mcl{D}}) q_{\text{ep}}(y_{\mbf{x}}|\mbf{y}_{\mcl{D}}, \mbf{x}^{\star})$. Hence, it is amenable to stochastic optimization like the optimization of $\alpha_{\text{sp}}^\star(\mbf{y}_{\mcl{D}}, \mbf{x})$. 

Given $\mcl{X}^\star$, the time complexity of evaluating $\alpha_{\text{ep}}^\star(\mbf{y}_{\mcl{D}}, \mbf{x})$ is mainly due to the EP procedure which incurs $\mcl{O}(n_{\text{ep}} |\mcl{X}^\star| |\mcl{D}|^3)$ where $n_{\text{ep}}$ is the number of EP iterations.
\begin{figure}[t]
\centering
\begin{tabular}{@{}cc@{}}	
	\includegraphics[height=0.22\textwidth]{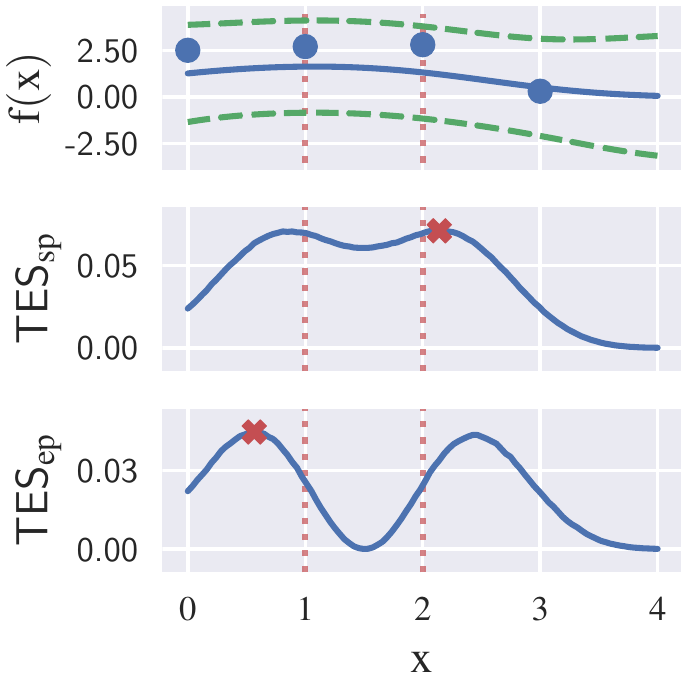}
	&
	\includegraphics[height=0.22\textwidth]{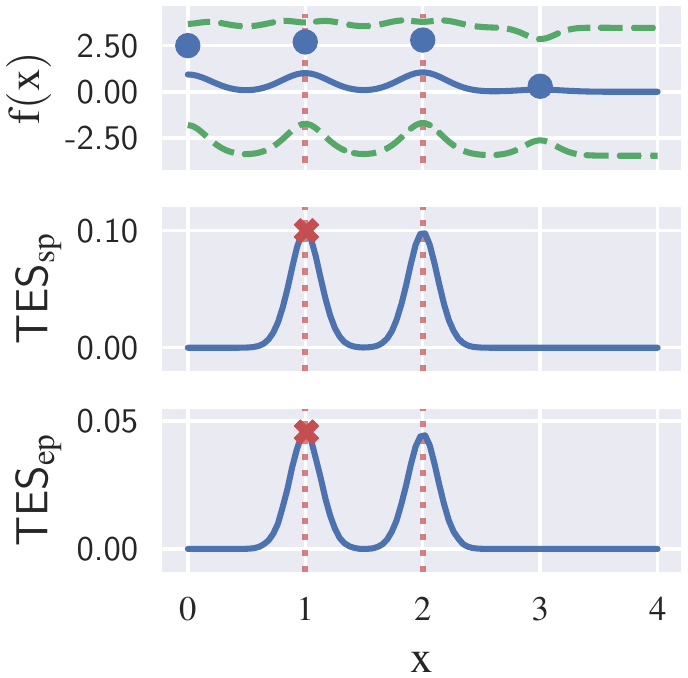}
	\\
	(a) Correlated $\mbf{f}_{\mcl{X}^{\star}}$
	&
	(b) Uncorrelated $\mbf{f}_{\mcl{X}^{\star}}$
\end{tabular}
\caption{TES\textsubscript{sp} and TES\textsubscript{ep} for (a) correlated $\mbf{f}_{\mcl{X}^{\star}}$ by varying the length-scale and (b) uncorrelated $\mbf{f}_{\mcl{X}^{\star}}$. The top plots show the GP posterior mean as a solid blue line, uncertainty (GP posterior variance) as dashed green lines, and data points as blue points. The dotted red lines show the input positions of $\mcl{X}^{\star}$. The red crosses indicate the maximizers of acquisition functions.}
\label{fig:compmpone}
\end{figure}

Fig.~\ref{fig:compmpone}b shows different acquisition functions in a synthetic example. Recall that TES measures the information gain on $\mbf{x}^{\star} \in \mcl{X}^{\star}$ by selecting the next input query $\mbf{x}\in\mcl{X}$ for evaluating $f$. So, for any $\mbf{x}$ that is far from $\mcl{X}^{\star}$, $f(\mbf{x})$ has little correlation with $\mbf{f}_{\mcl{X}^{\star}}$ and its TES value should be close to $0$. This observation holds for TES\textsubscript{sp} and TES\textsubscript{ep} in Fig.~\ref{fig:compmpone}b. 
Furthermore, TES\textsubscript{sp} and TES\textsubscript{ep} select an input in $\mcl{X}^{\star}$ when $\mbf{f}_{\mcl{X}^{\star}}$ are not strongly correlated, as shown in Fig.~\ref{fig:compmpone}b; red crosses are input queries $\mbf{x}$ that optimize the acquisition functions.
When there is a strong correlation within $\mbf{f}_{\mcl{X}^{\star}}$, TES\textsubscript{sp} and TES\textsubscript{ep} can exploit the correlation to select an informative input query $\mbf{x}$ around $\mcl{X}^{\star}$, as shown in Fig.~\ref{fig:compmpone}a.
As a result, to optimize TES, we should initialize the optimization with inputs in $\mcl{X}^{\star}$.
Another observation is that TES\textsubscript{ep} exploits more than TES\textsubscript{sp}, as shown in Appendix~\ref{app:epexploit}. 

\subsection{Generalization to Batch BO}
\label{bbo}
%
\begin{figure}[t]
\centering
\begin{tabular}{@{}cc@{}}
    \includegraphics[height=0.18\textwidth]{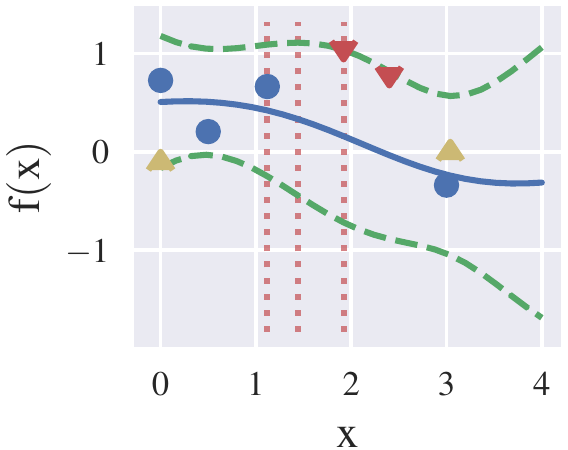}
     &
    \includegraphics[height=0.18\textwidth]{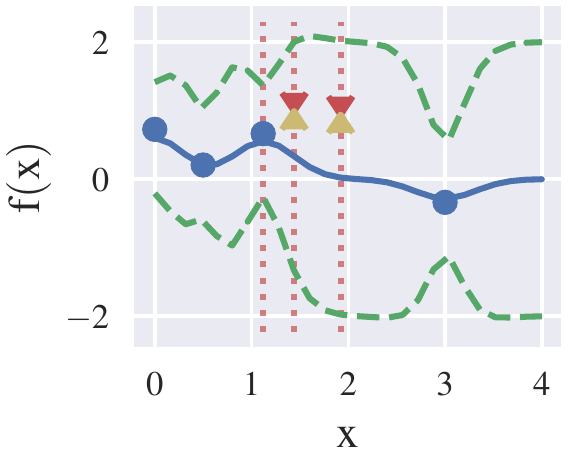}
     \\
     \\
    \includegraphics[height=0.22\textwidth]{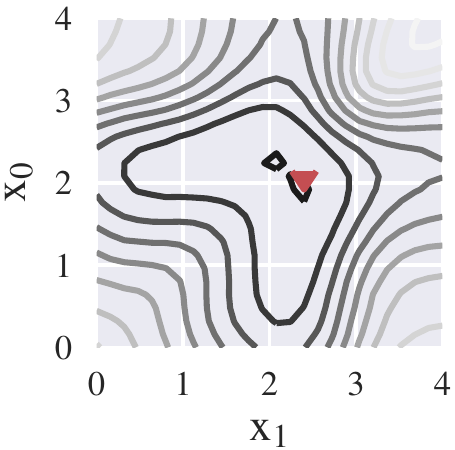}
     &
    \includegraphics[height=0.22\textwidth]{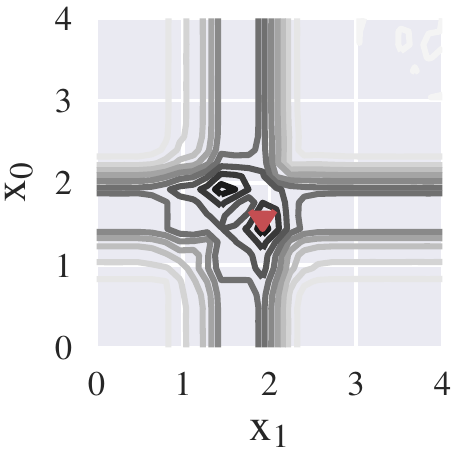}
     \\
     \\
    \includegraphics[height=0.22\textwidth]{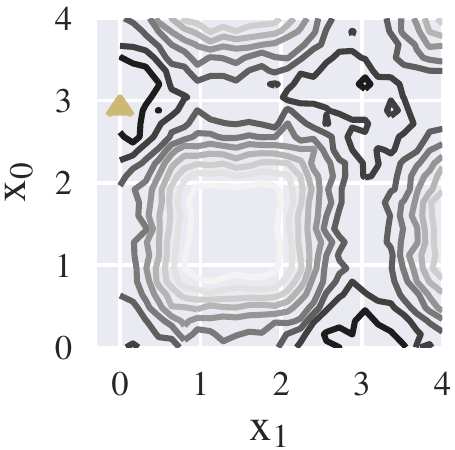}
     &
    \includegraphics[height=0.22\textwidth]{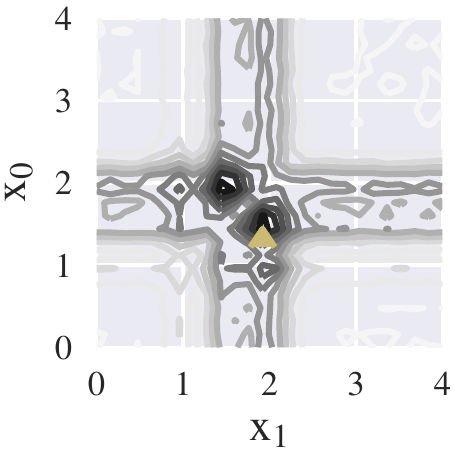}
\end{tabular}
\caption{TES\textsubscript{sp} and TES\textsubscript{ep} with batches $\mcl{B} = \{x_{0}, x_{1}\}$ for correlated $\mcl{X}^{\star}$ (left column) and uncorrelated $\mcl{X}^{\star}$ (right column) by varying the length-scale. The top, middle, and bottom plots are GP posteriors, TES\textsubscript{sp}, and TES\textsubscript{ep}, respectively. The notations are similar to Fig.~\ref{fig:compmpone}. Yellow caret-ups and red caret-downs denote the input query $\mbf{x}$ for TES\textsubscript{ep} and TES\textsubscript{sp}, respectively. The bottom $4$ plots are not perfectly symmetrical about $x_{0} = x_{1}$ due to sampling.}
\label{fig:batchex}
\end{figure}
%
In batch BO where a batch of inputs, denoted as $\mcl{B}$, is selected at each BO iteration, TES is defined as the amount of information gain about $\mbf{x}^{\star}$ through observing the noisy observations $\mbf{y}_{\mcl{B}}$ of the batch $\mcl{B}$:
\begin{equation}
\alpha^{\star}(\mbf{y}_{\mcl{D}}, \mcl{B}) = H(\mbf{y}_{\mcl{B}} | \mbf{y}_{\mcl{D}}) - \mbb{E}_{p(\mbf{x}^*| \mbf{y}_{\mcl{D}})} [ H(\mbf{y}_{\mcl{B}} | \mbf{y}_{\mcl{D}}, \mbf{x}^{\star})]\ .
\hspace{-2.2mm}
\label{eq:batchsoc}
\end{equation}
It can be approximated either stochastically with samples or deterministically with EP. 
The Gaussian predictive entropy $H(\mbf{y}_{\mcl{B}}| \mbf{y}_{\mcl{D}})$ naturally encourages the diversity of a batch, e.g., sampling of highly correlated observations in a batch (i.e., redundant information) is discouraged. This is different from several existing works such as BBO-LP \citep{gonzalez16batch} which requires a local penalization term based on the objective function's Lipschitz constant in addition to the acquisition function.
Increasing the batch size from one (in the previous section) to multiple input queries changes neither the set of samples $\mcl{F}_{\mbf{x}^{\star}}$ approximating $p(\mbf{f}_{\mcl{X}^{\star}}| \mbf{y}_{\mcl{D}}, \mbf{x}^{\star})$ in the stochastic approximation nor the approximate posterior $q_{\text{ep}}(\mbf{f}_{\mcl{X}^{\star}}|\mbf{y}_{\mcl{D}},\mbf{x}^{\star})$ for $p(\mbf{f}_{\mcl{X}^{\star}} | \mbf{y}_{\mcl{D}}, \mbf{x}^{\star})$ in the deterministic approximation.
Both approximations are applied in the same manner as those in the previous section (i.e., only once prior to the acquisition optimization in each BO iteration).
The only difference in batch BO for $|\mcl{B}| > 1$ is that $\mbf{y}_{\mcl{B}}$ is a multivariate r.v., so $q_{\text{sp}}(\mbf{y}_{\mcl{B}} | \mbf{y}_{\mcl{D}}, \mbf{x}^{\star})$ and $q_{\text{ep}}(\mbf{y}_{\mcl{B}} | \mbf{y}_{\mcl{D}}, \mbf{x}^{\star})$ are a mixture of multivariate Gaussian distributions and a multivariate Gaussian distribution, respectively. Still, samples of $\mbf{y}_{\mcl{B}}$ from these distributions can be drawn efficiently for the stochastic optimization of the acquisition function.

Batch TES with more than one input in the batch still exhibits similar behavior to that with only an input in the batch. 
That is, when $\mbf{f}_{\mcl{X}^{\star}}$ are not strongly correlated, a sensible strategy is to sample observations at inputs in $\mcl{X}^{\star}$ to identify $\mbf{x}^{\star}$ (the right column of Fig.~\ref{fig:batchex}). In contrast, when there is a strong correlation in $\mbf{f}_{\mcl{X}^{\star}}$, TES can exploit the correlation to sample informative observations at inputs around $\mcl{X}^{\star}$ (the left column of Fig.~\ref{fig:batchex}). Based on this observation, we also initialize the acquisition optimization with a batch of inputs in $\mcl{X}^\star$.

Note that $\alpha^{\star}(\mbf{y}_{\mcl{D}}, \mcl{B})$ is the information gain about $\mbf{x}^{\star}$ through observing $\mbf{y}_{\mcl{B}}$. Thus, it is comparable between batches of different sizes. As illustrated in Appendix~\ref{app:diffbatchsize}, the size of $\mcl{X}^\star$ should be at least the batch size to make the most out of the observations $\mbf{y}_{\mcl{B}}$. Besides, we can select the batch size adaptively at each BO iteration based on a trade-off between the increase in the information gain and the cost of an observation.

%
\section{Experiments}

Several existing acquisition functions and TES are evaluated empirically on synthetic functions and real-world optimization problems. We first show the competitive performance of TES on simple synthetic functions in Sections \ref{sec:B=1} and~\ref{sec:B>1}. Subsequently in Section \ref{sec:real-world}, we demonstrate that it can keep up the good performance in high-dimensional complex real-world problems.

The predictive performance of BO algorithms is measured by the \emph{immediate regret} (IR), which is defined as $f^* - f(\bar{\mbf{x}})$ where $\bar{\mbf{x}}$ is the maximizer of the posterior mean given $\mbf{y}_{\mcl{D}}$, and $f^*$ is the true global maximum of the objective function. 
To account for the randomness in the optimization and the sampling of noisy observations, experiments are repeated over several random runs. The natural logarithm of the average of IR over these random runs is reported. The box plots of IR are shown in Appendix~\ref{app:experiment}.
All experiments are initialized with a training set of $2$ data samples unless otherwise specified. 
These synthetic and the real-world functions are each modeled as a sample of a GP whose kernel hyperparameters 
are learned using maximum likelihood estimation \citep{rasmussen06}.
Adam \citep{kingma15adam} is used to optimize TES with $300$ iterations. At each stochastic iteration of Adam, $1000$ samples of $y_{\mbf{x}}$ (or $\mbf{y}_{\mcl{B}}$) are drawn. The experiments are performed on an Intel Xeon E5-2683 v4 CPU and an NVIDIA GTX 1080 GPU.

\subsection{Batch BO with $|\mcl{B}| = 1$}\label{sec:B=1}

We first consider BO algorithms that select $1$ input query in each BO iteration. We compare TES\textsubscript{ep} and TES\textsubscript{sp} with existing algorithms: EI \citep{movckus78}, UCB \citep{srinivas10ucb}, MES \citep{wang17mes}, and PES \citep{pes}. There are $5$ inputs in $\mcl{X}^{\star}$.

\begin{figure}[ht]
\centering
\begin{tabular}{@{}cc@{}}	
	\includegraphics[height=0.21\textwidth]{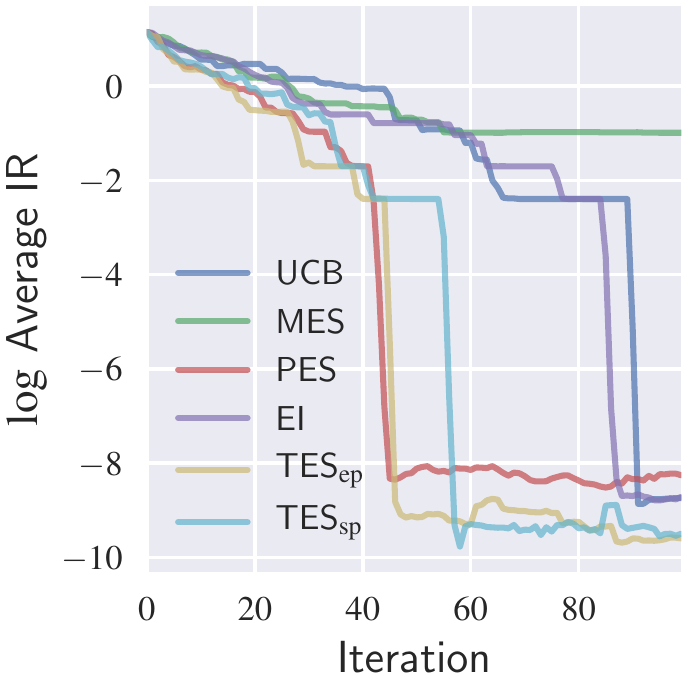}
	&
	\includegraphics[height=0.21\textwidth]{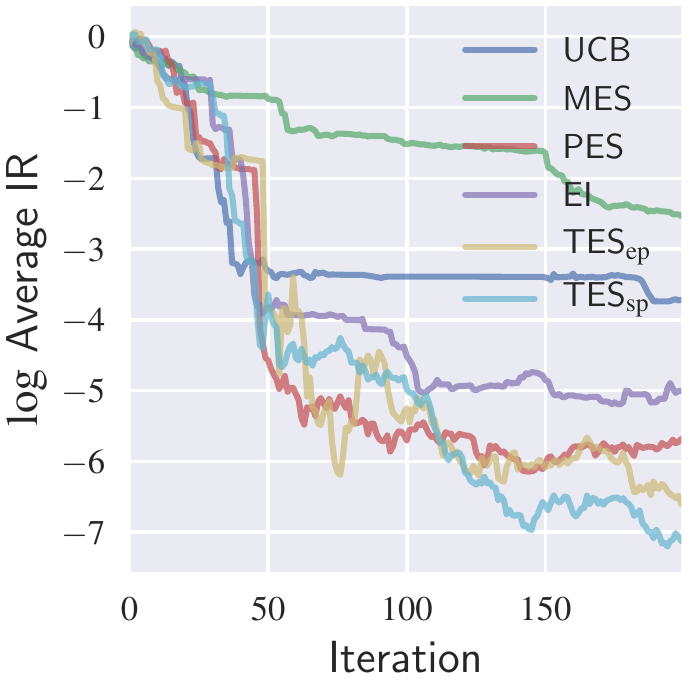}
	\\
	(a) The GP sampled function.
	&
	(b) log10P.
\end{tabular}
\caption{BO with $|\mcl{B}| = 1$.}
\label{fig:onebo}
\end{figure}

In the first experiment, we draw a function sample from a GP defined in $[0,10]^2$ with hyperparameters $\sigma_s^2 = 2$, $l_1 = l_2 = 1$, which is named the \emph{GP sampled function}. The noise variance to generate observations is $\sigma_n^2 = 10^{-4}$. The average results over $10$ random runs are shown in Fig.~\ref{fig:onebo}a.: TES\textsubscript{sp} and TES\textsubscript{ep} outperform other acquisition functions by converging to a better global maximum. Surprisingly, TES\textsubscript{ep} converges faster than TES\textsubscript{sp}. It could be because it exploits more than TES\textsubscript{sp} as mentioned above (illustrated in Appendix~\ref{app:epexploit}). The performance of PES closely matches those of TES acquisition functions, while it takes longer for UCB and EI to converge. MES does not perform as well as other acquisition functions that concentrate on the information gain of the maximizers (TES and PES). Experimental results on Branin and Hartmann-3D \citep{lizotte08} are in Appendix~\ref{app:experiment}.


The \emph{log10P} is a function that models the phosphorus field of Broom's Barn farm \citep{webster07} spatially distributed over a $1200$m by $680$m region discretized into a $31 \times 18$ grid of sampling locations. The logarithm to the base $10$ values of the amount of phosphorus are recorded. Fig.~\ref{fig:onebo}b shows the average results over $5$ random runs for the log10P with the input rescaled to $[0,1]^2$. To generate a continuous function from the dataset, we fit a GP on the dataset. This results in a GP posterior with the dataset as training data. 
The hyperparameters are $\sigma_s^2 = 7.669\times 10^{-2}$, $l_1 = 9.106\times 10^{-2}$, $l_2 = 3.971\times 10^{-2}$, and $\sigma_n^2 = 2.511\times 10^{-2}$. 
The posterior mean of this GP 
is taken as the objective function to optimize. Note that the noise variance in this experiment is larger than that of the previous experiments. 
Hence, it is advantageous for information-based acquisition functions such as TES and PES as they tend to explore more than the others.
Fig.~\ref{fig:onebo}b shows that TES\textsubscript{sp} converges to a better maximum than others. 
%
\subsection{Batch BO with $|\mcl{B}| > 1$}\label{sec:B>1}

In this section, we consider BO of batch size larger than~$1$.
We compare our algorithms with QEI-CL \citep{ginsbourger10}, BUCB \citep{desautels14}, QEI \citep{marmin15}, and LP-EI, which is the BBO-LP \citep{gonzalez16batch} by plugging in EI. 

\begin{figure}
\centering 
\begin{tabular}{@{}ccc@{}}	
	\includegraphics[height=0.15\textwidth]{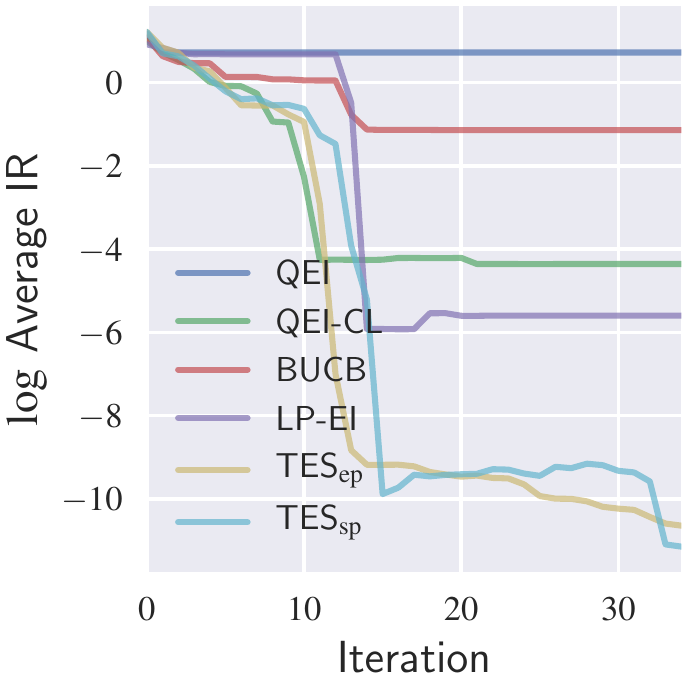}
    &
	\includegraphics[height=0.15\textwidth]{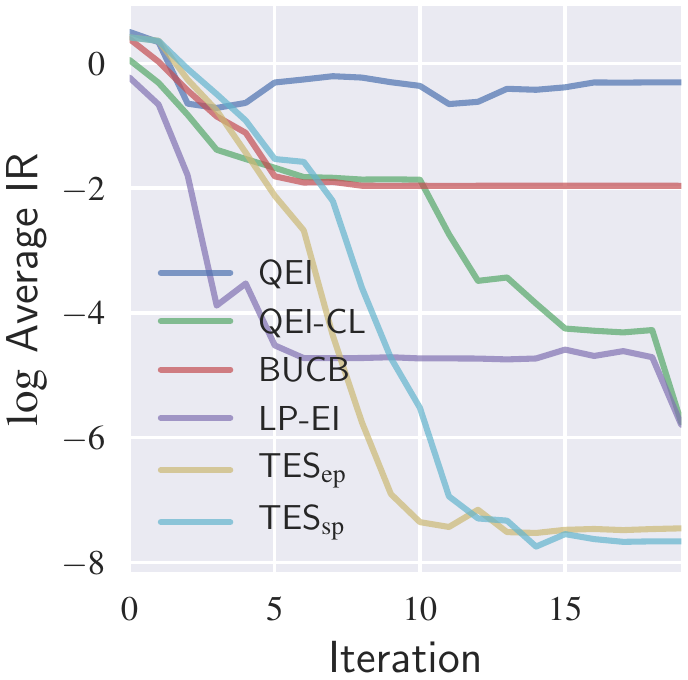}
	&
	\includegraphics[height=0.15\textwidth]{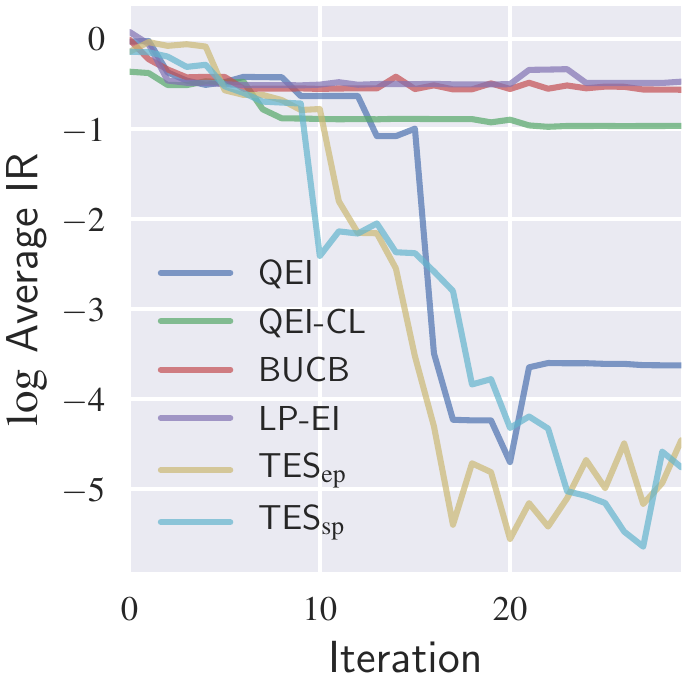}
	\\
	(a)
	&
	(b)
	&
	(c)
	\\
	\includegraphics[height=0.15\textwidth]{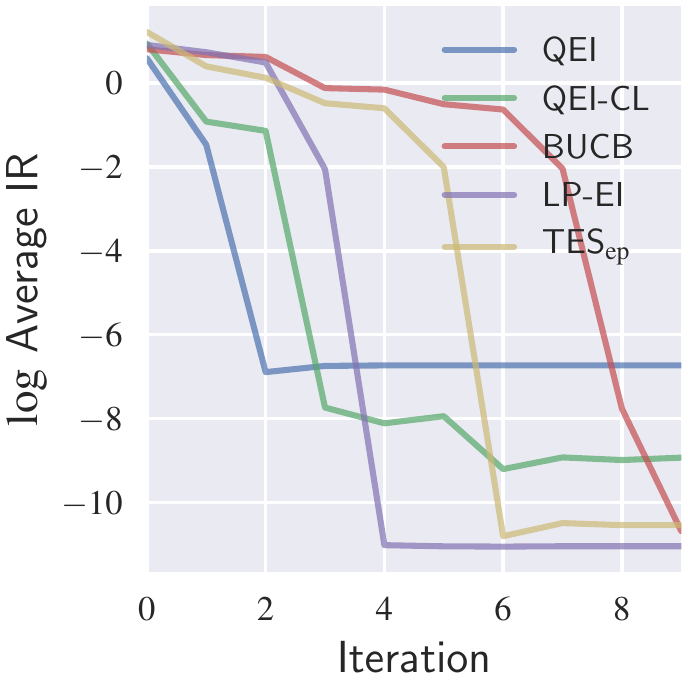}
	&
	\includegraphics[height=0.15\textwidth]{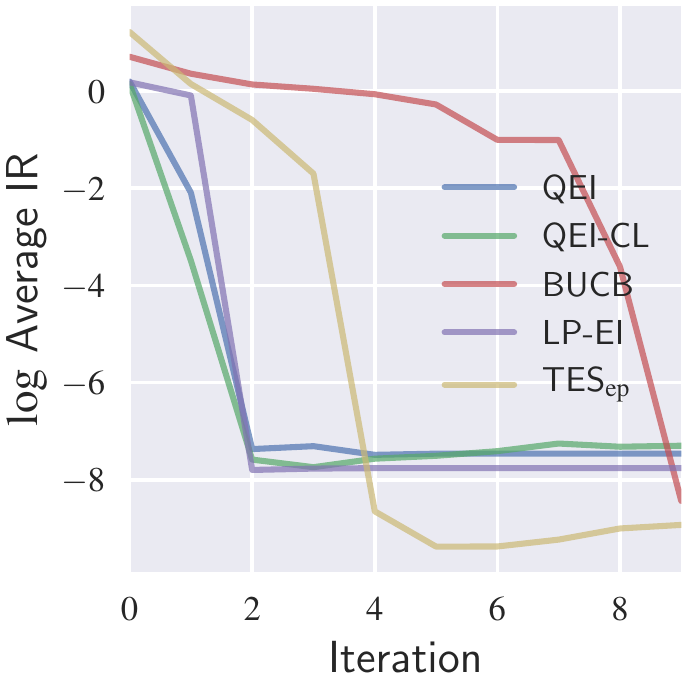}
	&
	\includegraphics[height=0.15\textwidth]{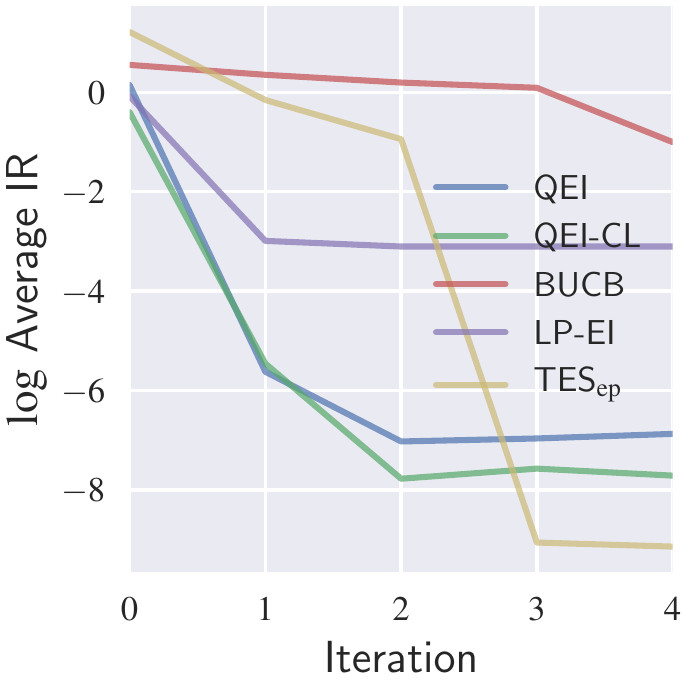}
	\\
	(d)
	&
	(e)
	&
	(f)
	\\
	\includegraphics[height=0.15\textwidth]{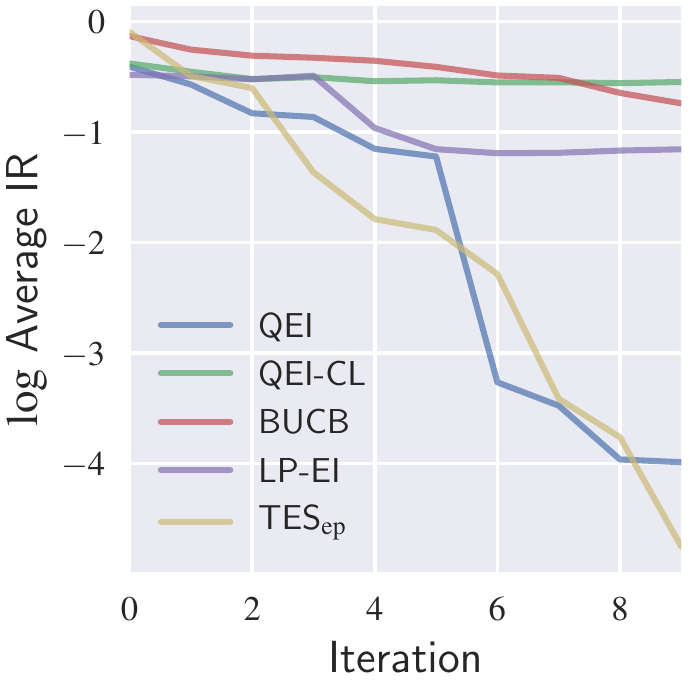}
	&
	\includegraphics[height=0.15\textwidth]{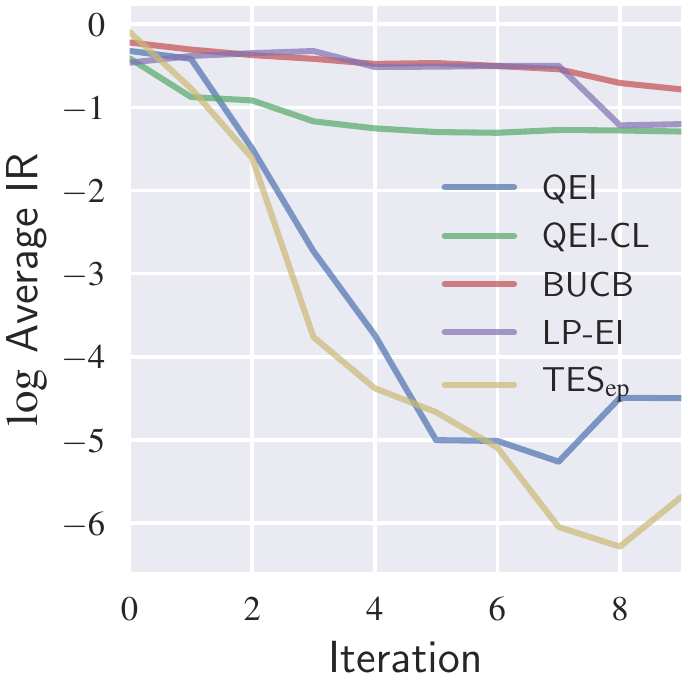}
	&
	\includegraphics[height=0.15\textwidth]{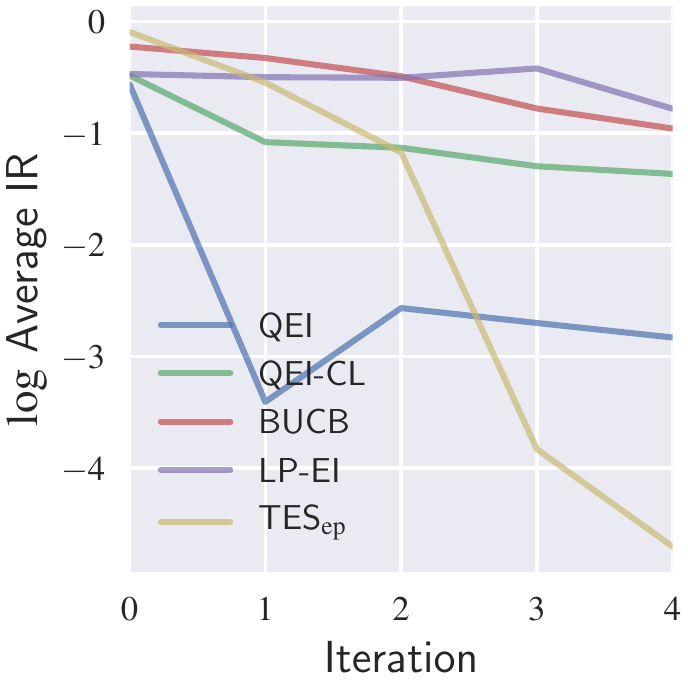}
	\\
	(g)
	&
	(h)
	&
	(i)
\end{tabular}
\caption{Batch BO: $|\mcl{B}| = 3$ on (a) the GP sampled function, (b) Hartmann-4D, and (c) the log10P;
$\mcl{B}$ of size (d) 10, (e) 20, and (f) 30 on the GP sampled function; and $\mcl{B}$ of size (g) 10, (h) 20, and (i) 40 on the log10P.}
\label{fig:bobatch}
\end{figure}

Figs.~\ref{fig:bobatch}a-c shows BO results with $|\mcl{B}| = 3$ 
over $3$ random experiments
for the GP sampled function, Hartmann-4D, and log10P. 
The size of $\mcl{X}^\star$ is set to $5$.
Overall, both TES\textsubscript{sp} and TES\textsubscript{ep} achieve good performance by converging to good maxima. 
The difference in the performance can be explained by the fact that TES can exploit the maximizer samples from the GP posterior to quickly explore the function domain.

To illustrate the scalability of TES\textsubscript{ep}, we experiment with large batch sizes: $|\mcl{B}| = 10$, $20$ and $30$ for the GP sampled function, and $|\mcl{B}| = 10$, $20$ and $40$ for the log10P in Figs.~\ref{fig:bobatch}d-i.
The size of $\mcl{X}^\star$ is set to be the same as that of $\mcl{B}$. It is to make the most of the information in the batch as explained in Appendix~\ref{app:diffbatchsize}.
In the GP sampled function experiments (Figs.~\ref{fig:bobatch}d-f), though other acquisition functions converge faster than TES\textsubscript{ep}, they seem to be stuck in suboptimal maxima except for LP-EI in Fig.~\ref{fig:bobatch}d. At the initial iterations with little observations, $\mcl{X}^\star$ are likely to scatter over the whole input space, so TES\textsubscript{ep} explores more than other acquisition functions. At the latter iterations with a sufficient number of observations, TES can exploit to discover a good maximum. 
In the log10P experiments with noisier observations (Figs.~\ref{fig:bobatch}g-i), we observe that only QEI shows similar performance to our TES\textsubscript{ep}, though our TES\textsubscript{ep} outperforms for batch sizes of $20$ and $40$. It could be because both QEI and TES\textsubscript{ep} jointly optimize the batch, which is an advantage compared with other greedy batch acquisition functions in these experiments. Another observation for TES\textsubscript{ep} in Figs.~\ref{fig:bobatch}d-i is that increasing the batch size improves the efficiency of BO with respect to the number of iterations.

\setlength{\tabcolsep}{3pt}

\subsection{Optimizing Complex Real-World Problems}\label{sec:real-world}

In this section, we aim to illustrate that TES works well in real life. We apply our more scalable TES\textsubscript{ep} algorithm to two real-world scenarios that can be regarded as black-box optimization problems. Other batch BO acquisition functions mentioned previously are evaluated as comparisons.

The first real-world experiment is motivated by the computationally expensive procedure of neural architecture search in convolutional neural networks (CNNs). We train a 2-layer CNN on the CIFAR-10 dataset and aim to efficiently search for the best combination of 7 hyperparameters: batch size, number of convolutional filters, number of dense units, dropout rate, L2 regularizer, learning rate, and learning rate decay. The bounds are $[16, 512]$, $[6, 256]$, $[128, 2048]$, $[0, 0.75]$, $[10^{-7}, 10^{-3}]$, $[10^{-7}, 10^{-1}]$, $[10^{-7}, 10^{-3}]$ respectively and the first three hyperparameters are integers.
The IR is calculated by $1$ minus the test accuracy.


\begin{figure}
\centering
\begin{tabular}{@{}cc@{}}	
	\includegraphics[height=0.2\textwidth]{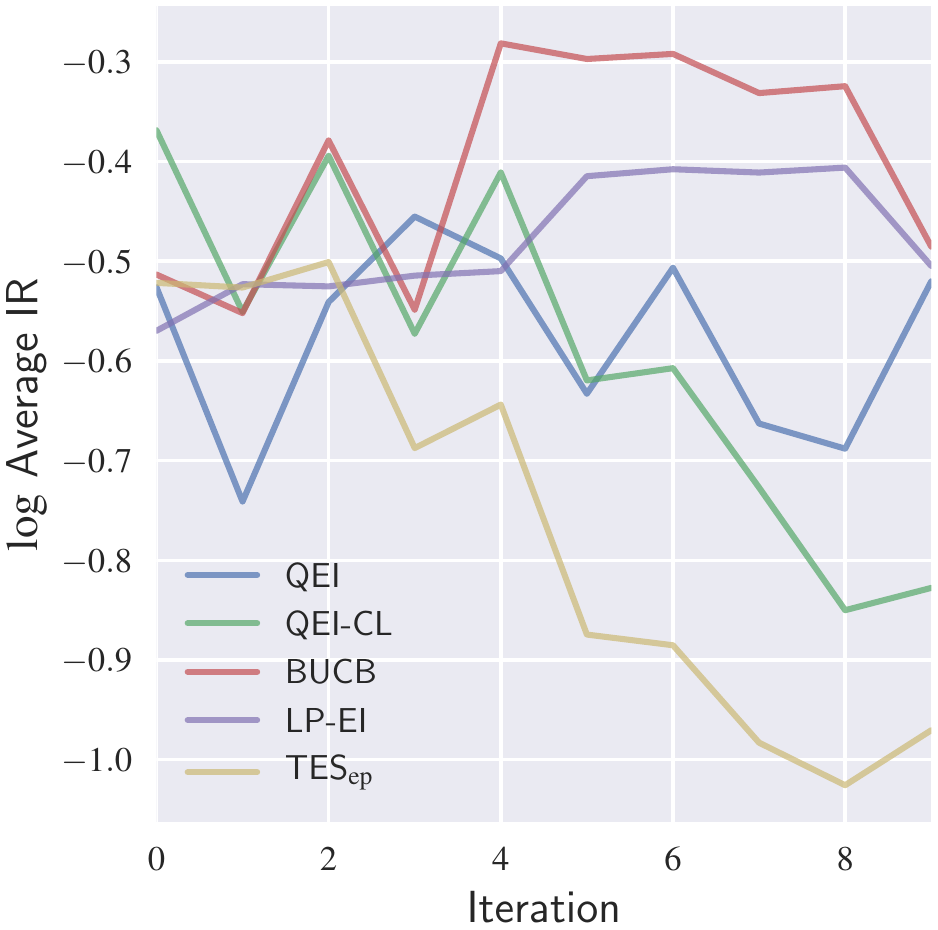}
	&
	\includegraphics[height=0.2\textwidth]{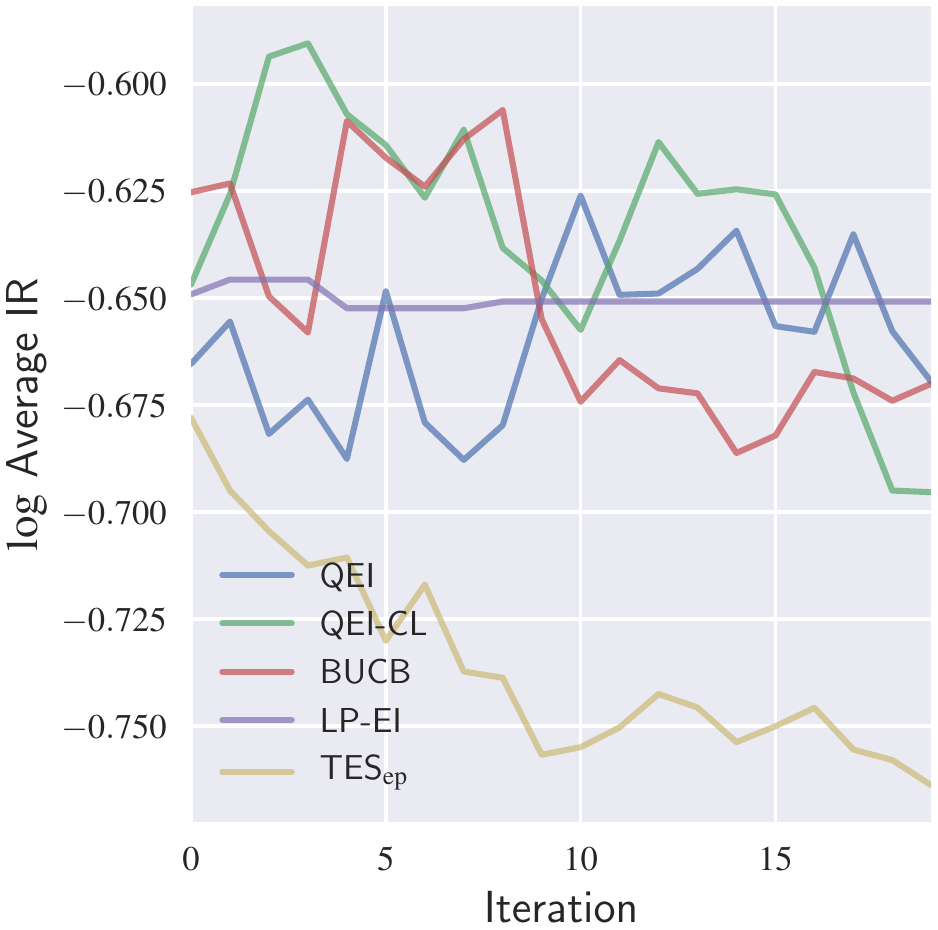}
	\\
	(a) The CIFAR-10.
	&
	(b) The face attack function.
\end{tabular}
\caption{BO results on real-world problems.}
\label{fig:real-world}
\end{figure}

Fig. \ref{fig:real-world}a shows BO results with  $|\mcl{B}| = 20$ over $5$ random experiments for CIFAR-10. The initial training set consists of $10$ data samples. We observe greater fluctuations in the regret for all methods because the outputs of complex problems generally contain larger noises than those of synthetic functions. Overall, TES\textsubscript{ep} still converges the fastest to the lowest regret. All other methods except QEI-CL are stuck in suboptimal maxima obtained at the beginning few iterations. This empirical experiment shows that TES\textsubscript{ep} can explore multi-dimensional search spaces well in real life and find the best maximum among all the methods tested.

In the second real-world scenario, we synthesize `physically realizable' faces to fool a black-box \emph{face recognition system} (FRS) \citep{saha20face}. We formalize the problem as an optimization task, illustrating the potential privacy and security risks of such systems under malicious attacks.
More precisely, by altering some parameters representing specific facial features, the attacker's face is synthesized such that it minimizes the distance $d$ to a victim's face perceived by an FRS. The victim could be any person registered in the FRS, while the attacker is not. The lower the distance, the more likely the attacker could successfully fool and bypass the security of the system. Therefore, the performance metric is the distance $d$. 

\begin{figure}[t!]
    \centering
    \begin{tabular}{@{}ccc@{}}
    	\includegraphics[ height=0.17\textwidth]{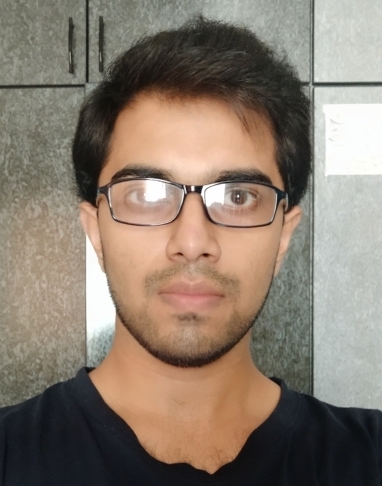}
    	&
        \includegraphics[height=0.17\textwidth]{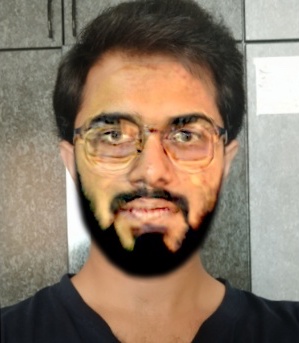}
    	&
    	\includegraphics[ height=0.17\textwidth]{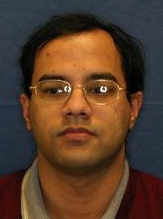}
    	\\
        (a)
    	&
    	(b)
    	&
    	(c)
    	\\
    \end{tabular}
    \caption{Visualization of a typical result. (a) is the original image of the attacker. (b) shows the synthesized image of the attacker. (c) is the `matched' image in the gallery.}
    \label{fig:matched_face}
\end{figure}

In our experiments, we use the Python \emph{face\_recognition} library \citep{facereg-website} as the target black-box FRS and alter 8 parameters linked to facial features such as facial hair (e.g., beard, mustache), marks (e.g., forehead scar, cheek scar, mole), eyeglasses (e.g., squared, round) and expressions (e.g., smile). Note that bounds are imposed on the facial feature parameters to ensure more realistic appearances of the synthesized faces, i.e., `physically realizable' faces. The multimodal discriminant analysis (MMDA) face synthesizer model described by~\citet{sim15face} is used. 
The library outputs a $128$-dimension face encoding for each input face image and considers two faces as a pair of `match' if the Euclidean distance between the two encodings is lower than a threshold $\theta = 0.5$.  Given a gallery of 41 face images selected from the CMU Multi-PIE dataset \citep{cmu-dataset} which consists of 36 males and 5 females, our objective is to modify an attacker's face in Fig.~\ref{fig:matched_face}a so that its distance to a registered face in the gallery is minimized. We hope the distance to go below $\theta$ for a successful break-in attack.

Fig. \ref{fig:real-world}b shows BO results with $|\mcl{B}| = 20$ over $5$ random experiments for the face attack function. The initial training size is again $10$. Similar to what we observed in the CIFAR-10 experiment, TES\textsubscript{ep} achieves the lowest regret while other methods only obtain suboptimal maxima. TES\textsubscript{ep} explores over the whole input space for about 10 iterations (or 200 samples) before turning to exploitation to discover a good maximum. Similar behavior is seen in Fig.~\ref{fig:bobatch}c. Additionally, the box plots of IR are included in Appendix \ref{app:experiment}.


We show the visualization of a typical result after iterations of optimization using BO with our TES\textsubscript{ep} acquisition function in Fig.~\ref{fig:matched_face}. In this particular case, the attacker's face is modified to Fig.~\ref{fig:matched_face}b such that the distance to a registered face encoding in the gallery is optimized to $0.4534$, implying that the privacy and security of the FRS have been compromised since the $d$ is smaller than the threshold $\theta=0.5$.
Notably, the modified face is `physically realizable', i.e., the attacker could grow a beard, change the glasses, put on some scars makeup and smile to fool the FRS.
The victim's face is shown in Fig.~\ref{fig:matched_face}c.
Therefore, our proposed algorithm is effective in such real-world optimization problems.

\section{Conclusion}
This paper presents \emph{trusted-maximizers entropy search} (TES) for performing Bayesian optimization and its batch variant.
Both stochastic approximation with sampling and deterministic approximation with a Gaussian distribution are investigated, which results in TES\textsubscript{sp} and TES\textsubscript{ep}, respectively. While TES\textsubscript{sp} produces a better approximation of the posterior belief, TES\textsubscript{ep} is less expensive to evaluate. Therefore, TES\textsubscript{ep} can be scaled to BO with large batch sizes.
TES\textsubscript{sp} and TES\textsubscript{ep} achieve favorable performance compared with existing acquisition functions 
in experiments with synthetic benchmark objective functions and complex real-world optimization problems.

\begin{acknowledgements} 
This research is supported by the National Research Foundation, Singapore under its Strategic Capability Research Centres Funding Initiative and the Singapore Ministry of Education Academic Research Fund Tier $1$. Any opinions, findings, and conclusions or recommendations expressed in this material are those of the author(s) and do not reflect the views of National Research Foundation, Singapore.

\end{acknowledgements}

\bibliography{bo}

\begin{thebibliography}{37}
\providecommand{\natexlab}[1]{#1}
\providecommand{\url}[1]{\texttt{#1}}
\expandafter\ifx\csname urlstyle\endcsname\relax
  \providecommand{\doi}[1]{doi: #1}\else
  \providecommand{\doi}{doi: \begingroup \urlstyle{rm}\Url}\fi

\bibitem[Brochu et~al.(2010{\natexlab{a}})Brochu, Brochu, and {de
  Freitas}]{brochu10animate}
E.~Brochu, T.~Brochu, and N.~{de Freitas}.
\newblock A {Bayesian} interactive optimization approach to procedural
  animation design.
\newblock In \emph{Proc. ACM SIGGRAPH/Eurographics Symposium on Computer
  Animation}, pages 103--112, 2010{\natexlab{a}}.

\bibitem[Brochu et~al.(2010{\natexlab{b}})Brochu, Cora, and {de
  Freitas}]{brochu10tut}
E.~Brochu, V.~M. Cora, and N.~{de Freitas}.
\newblock A tutorial on {Bayesian} optimization of expensive cost functions,
  with application to active user modeling and hierarchical reinforcement
  learning.
\newblock \emph{arXiv:1012.2599}, 2010{\natexlab{b}}.

\bibitem[Dai et~al.(2019)Dai, Yu, Low, and Jaillet]{dai2019bayesian}
Z.~Dai, H.~Yu, B.~K.~H. Low, and P.~Jaillet.
\newblock Bayesian optimization meets {Bayesian} optimal stopping.
\newblock In \emph{Proc. {ICML}}, pages 1496--1506, 2019.

\bibitem[Dai et~al.(2020)Dai, Chen, Low, Jaillet, and Ho]{dai2020r2}
Z.~Dai, Y.~Chen, B.~K.~H. Low, P.~Jaillet, and T.~Ho.
\newblock {R2-B2}: Recursive reasoning-based {Bayesian} optimization for
  no-regret learning in games.
\newblock In \emph{Proc. {ICML}}, 2020.

\bibitem[Daxberger and Low(2017)]{daxberger2017distributed}
E.~A. Daxberger and B.~K.~H. Low.
\newblock Distributed batch {Gaussian} process optimization.
\newblock In \emph{Proc. {ICML}}, pages 951--960, 2017.

\bibitem[Desautels et~al.(2014)Desautels, Krause, and Burdick]{desautels14}
T.~Desautels, A.~Krause, and J.~Burdick.
\newblock Parallelizing exploration-exploitation tradeoffs in {Gaussian}
  process bandit optimization.
\newblock \emph{JMLR}, 15\penalty0 (1):\penalty0 3873--3923, 2014.

\bibitem[Geitgey(2017)]{facereg-website}
A.~Geitgey.
\newblock Face recognition, 2017.
\newblock URL \url{https://github.com/ageitgey/face_recognition}.

\bibitem[Ginsbourger et~al.(2010)Ginsbourger, {Le Riche}, and
  Carraro]{ginsbourger10}
D.~Ginsbourger, R.~{Le Riche}, and L.~Carraro.
\newblock Kriging is well-suited to parallelize optimization.
\newblock In \emph{Computational intelligence in expensive optimization
  problems}, pages 131--162. Springer, 2010.

\bibitem[Gonz{\'a}lez et~al.(2016)Gonz{\'a}lez, Dai, Hennig, and
  Lawrence]{gonzalez16batch}
J.~Gonz{\'a}lez, Z.~Dai, P.~Hennig, and N.~Lawrence.
\newblock Batch {Bayesian} optimization via local penalization.
\newblock In \emph{Proc. AISTATS}, pages 648--657, 2016.

\bibitem[Gross et~al.(2010)Gross, Matthews, Cohn, Kanade, and
  Baker]{cmu-dataset}
Ralph Gross, Iain Matthews, Jeffrey Cohn, Takeo Kanade, and Simon Baker.
\newblock Multi-pie.
\newblock \emph{Image and vision computing}, 28\penalty0 (5):\penalty0
  807--813, 2010.

\bibitem[Hennig and Schuler(2012)]{es}
P.~Hennig and C.~J. Schuler.
\newblock Entropy search for information-efficient global optimization.
\newblock \emph{JMLR}, pages 1809--1837, 2012.

\bibitem[{Hern{\'a}ndez-Lobato} et~al.(2014){Hern{\'a}ndez-Lobato}, Hoffman,
  and Ghahramani]{pes}
J.~M. {Hern{\'a}ndez-Lobato}, M.~W. Hoffman, and Z.~Ghahramani.
\newblock Predictive entropy search for efficient global optimization of
  black-box functions.
\newblock In \emph{Proc. {NIPS}}, pages 918--926, 2014.

\bibitem[Hoang et~al.(2018)Hoang, Hoang, Ouyang, and Low]{NghiaAAAI18}
T.~N. Hoang, Q.~M. Hoang, R.~Ouyang, and B.~K.~H. Low.
\newblock Decentralized high-dimensional {Bayesian} optimization with factor
  graphs.
\newblock In \emph{Proc. {AAAI}}, pages 3231--3238, 2018.

\bibitem[Hoffman and Ghahramani(2015)]{hoffman15opes}
M.~W. Hoffman and Z.~Ghahramani.
\newblock Output-space predictive entropy search for flexible global
  optimization.
\newblock In \emph{NIPS workshop on Bayesian Optimization}, 2015.

\bibitem[Kharkovskii et~al.(2020)Kharkovskii, Dai, and
  Low]{kharkovskii2020private}
D.~Kharkovskii, Z.~Dai, and B.~K.~H. Low.
\newblock Private outsourced {Bayesian} optimization.
\newblock In \emph{Proc. {ICML}}, 2020.

\bibitem[Kingma and Ba(2015)]{kingma15adam}
D.~P. Kingma and J.~Ba.
\newblock Adam: {A} method for stochastic optimization.
\newblock In \emph{Proc. ICLR}, 2015.

\bibitem[Kingma and Welling(2013)]{kingma2013auto}
D.~P. Kingma and M.~Welling.
\newblock Auto-encoding variational {B}ayes.
\newblock \emph{arXiv:1312.6114}, 2013.

\bibitem[Kushner(1964)]{kushner1964new}
H.~J. Kushner.
\newblock A new method of locating the maximum point of an arbitrary multipeak
  curve in the presence of noise.
\newblock \emph{Journal of basic engineering}, 86\penalty0 (1):\penalty0
  97--106, 1964.

\bibitem[Ling et~al.(2016)Ling, Low, and Jaillet]{ling16}
C.~K. Ling, B.~K.~H. Low, and P.~Jaillet.
\newblock {Gaussian} process planning with {Lipschitz} continuous reward
  functions: Towards unifying {Bayesian} optimization, active learning, and
  beyond.
\newblock In \emph{Proc. {AAAI}}, pages 1860--1866, 2016.

\bibitem[Lizotte(2008)]{lizotte08}
D.~J. Lizotte.
\newblock \emph{Practical {Bayesian} optimization}.
\newblock PhD thesis, University of {Alberta}, Canada, 2008.

\bibitem[Marmin et~al.(2015)Marmin, Chevalier, and Ginsbourger]{marmin15}
S.~Marmin, C.~Chevalier, and D.~Ginsbourger.
\newblock Differentiating the multipoint expected improvement for optimal batch
  design.
\newblock In \emph{International Workshop on Machine Learning, Optimization and
  Big Data}, pages 37--48. Springer, 2015.

\bibitem[Minka(2001)]{minka01ep}
T.~P. Minka.
\newblock \emph{A family of algorithms for approximate {Bayesian} inference}.
\newblock PhD thesis, Massachusetts Institute of Technology, 2001.

\bibitem[Mockus et~al.(1978)Mockus, Tie{\v{s}}is, and
  {\v{Z}}ilinskas]{movckus78}
J.~Mockus, V.~Tie{\v{s}}is, and A.~{\v{Z}}ilinskas.
\newblock The application of {B}ayesian methods for seeking the extremum.
\newblock In L.~C.~W. Dixon and G.~P. Szeg{\"{o}}, editors, \emph{Towards
  Global Optimization 2}, pages 117--129. North-Holland Publishing Company,
  1978.

\bibitem[Nguyen et~al.(2021{\natexlab{a}})Nguyen, Low, and
  Jaillet]{Nguyen21active}
Q.~P. Nguyen, B.~K.~H. Low, and P.~Jaillet.
\newblock An information-theoretic framework for unifying active learning
  problems.
\newblock In \emph{Proc. {AAAI}}, pages 9126--9134, 2021{\natexlab{a}}.

\bibitem[Nguyen et~al.(2021{\natexlab{b}})Nguyen, Tay, Low, and
  Jaillet]{Nguyen21topk}
Q.~P. Nguyen, S.~Tay, B.~K.~H. Low, and P.~Jaillet.
\newblock Top-k ranking {Bayesian} optimization.
\newblock In \emph{Proc. {AAAI}}, pages 9135--9143, 2021{\natexlab{b}}.

\bibitem[Rasmussen and Williams(2006)]{rasmussen06}
C.~E. Rasmussen and C.~K.~I. Williams.
\newblock \emph{Gaussian processes for machine learning}.
\newblock MIT Press, 2006.

\bibitem[Ru et~al.(2018)Ru, McLeod, Granziol, and Osborne]{ru2018fast}
B.~Ru, M.~McLeod, D.~Granziol, and M.~A. Osborne.
\newblock Fast information-theoretic {Bayesian} optimisation.
\newblock In \emph{Proc. ICML}, pages 4381--4389, 2018.

\bibitem[{Saha} and {Sim}(2020)]{saha20face}
S.~{Saha} and T.~{Sim}.
\newblock Is face recognition safe from realizable attacks?
\newblock In \emph{International Joint Conference on Biometrics}, 2020.

\bibitem[Shah and Ghahramani(2015)]{ppes}
A.~Shah and Z.~Ghahramani.
\newblock Parallel predictive entropy search for batch global optimization of
  expensive objective functions.
\newblock In \emph{Proc. NIPS}, pages 3330--3338, 2015.

\bibitem[Shahriari et~al.(2015)Shahriari, Swersky, Wang, Adams, and {de
  Freitas}]{shahriari15}
B.~Shahriari, K.~Swersky, Z.~Wang, R.~Adams, and N.~{de Freitas}.
\newblock Taking the human out of the loop: A review of {Bayesian}
  optimization.
\newblock \emph{Proceedings of the {IEEE}}, 104\penalty0 (1):\penalty0
  148--175, 2015.

\bibitem[{Sim} and {Zhang}(2015)]{sim15face}
T.~{Sim} and L.~{Zhang}.
\newblock Controllable face privacy.
\newblock In \emph{2015 11th IEEE International Conference and Workshops on
  Automatic Face and Gesture Recognition (FG)}, volume~04, pages 1--8, 2015.

\bibitem[Snoek et~al.(2012)Snoek, Larochelle, and Adams]{snoek12}
J.~Snoek, H.~Larochelle, and R.~Adams.
\newblock Practical bayesian optimization of machine learning algorithms.
\newblock In \emph{Proc. NIPS}, pages 2951--2959, 2012.

\bibitem[Srinivas et~al.(2010)Srinivas, Krause, Kakade, and
  Seeger]{srinivas10ucb}
N.~Srinivas, A.~Krause, S.~Kakade, and M.~Seeger.
\newblock Gaussian process optimization in the bandit setting: {No} regret and
  experimental design.
\newblock In \emph{Proc. ICML}, pages 1015--1022, 2010.

\bibitem[Wang and Jegelka(2017)]{wang17mes}
Z.~Wang and S.~Jegelka.
\newblock Max-value entropy search for efficient {Bayesian} optimization.
\newblock In \emph{Proc. ICML}, pages 3627--3635, 2017.

\bibitem[Webster and Oliver(2007)]{webster07}
R.~Webster and M.~Oliver.
\newblock \emph{Geostatistics for environmental scientists}.
\newblock John Wiley \& Sons, 2007.

\bibitem[Zhang et~al.(2017)Zhang, Hoang, Low, and
  Kankanhalli]{zhang2017information}
Y.~Zhang, T.~N. Hoang, B.~K.~H. Low, and M.~Kankanhalli.
\newblock Information-based multi-fidelity {Bayesian} optimization.
\newblock In \emph{Proc. {NeurIPS} Workshop on {Bayesian} Optimization}, 2017.

\bibitem[Zhang et~al.(2019)Zhang, Dai, and Low]{zhang2019bayesian}
Y.~Zhang, Z.~Dai, and B.~K.~H. Low.
\newblock Bayesian optimization with binary auxiliary information.
\newblock In \emph{Proc. {UAI}}, pages 1222--1232, 2019.

\end{thebibliography}

\clearpage
\appendix

\section{Evaluation of $p(\mbf{x}^\star | \mbf{y}_{\mcl{D}})$}
\label{app:opmf}

The probability $p(\mbf{x}^{\star}|\mbf{y}_{\mcl{D}})$ can be expressed as:
\begin{align*}
 p(\mbf{x}^{\star}|\mbf{y}_{\mcl{D}})
	&= p(f(\mbf{x}^{\star}) \ge f(\mbf{x})\ \forall \mbf{x} \in \mcl{X}^{\star} | \mbf{y}_{\mcl{D}})\\
 	&= \displaystyle p(\mbf{J}_{\mbf{x}^{\star}} \mbf{f}_{\mcl{X}^{\star}} \le 0 | \mbf{y}_{\mcl{D}})
\end{align*}
where $\mbf{J}_{\mbf{x}^{\star}}$ is a matrix of size $|\mcl{X}^\star| \times |\mcl{X}^\star|$ with $0$ entries except for $\mbf{J}_{ii} = 1$, $\mbf{J}_{ij} = -1$ if $i \neq j$ where $j$ is the index of $\mbf{x}^{\star}$ in $\mcl{X}^{\star}$. As $p(\mbf{f}_{\mcl{X}^{\star}}|\mbf{y}_{\mcl{D}})$ is the p.d.f. of a multivariate Gaussian~\eqref{eq:gppost}, so is $p(\mbf{J}_{\mbf{x}^{\star}} \mbf{f}_{\mcl{X}^{\star}} | \mbf{y}_{\mcl{D}})$. Hence, $p(\mbf{J}_{\mbf{x}^{\star}} \mbf{f}_{\mcl{X}^{\star}} \le 0 | \mbf{y}_{\mcl{D}})$ can be computed efficiently.

\section{Importance Sampling from $p(\mbf{f}_{\mcl{X}^{\star}}| \mbf{y}_{\mcl{D}}, \mbf{x}^{\star})$}
\label{app:imppostcand}

Let $\mbf{f}_{\setminus \star}\triangleq (f(\mbf{x}'))^{\top}_{\mbf{x}' \in \mcl{X}^{\star}\setminus\{\mbf{x}^{\star}\}}$ denote the function values at $\mcl{X}^{\star} \setminus \{\mbf{x}^{\star}\}$. We have  $p(\mbf{f}_{\mcl{X}^{\star}}| \mbf{y}_{\mcl{D}}, \mbf{x}^{\star}) \propto p(\mbf{f}_{\mcl{X}^{\star}}, \mbf{x}^{\star} | \mbf{y}_{\mcl{D}})$ which equals
\begin{equation*}
p(\mbf{x}^{\star} | \mbf{f}_{\setminus \star}, \mbf{y}_{\mcl{D}})
p(\mbf{f}_{\setminus \star} | \mbf{y}_{\mcl{D}})
p(f(\mbf{x}^\star) | \mbf{x}^{\star}, \mbf{f}_{\setminus \star}, \mbf{y}_{\mcl{D}})\ .
\end{equation*}
We can first draw samples of $\mbf{f}_{\mcl{X}^{\star}}$ from the p.d.f. $p(\mbf{f}_{\setminus \star} | \mbf{y}_{\mcl{D}})
p(f(\mbf{x}^\star) | \mbf{x}^{\star}, \mbf{f}_{\setminus \star}, \mbf{y}_{\mcl{D}})$, then weight these samples with 
$p(\mbf{x}^{\star} | \mbf{f}_{\setminus \star}, \mbf{y}_{\mcl{D}})$.
Sampling from $p(\mbf{f}_{\setminus \star} | \mbf{y}_{\mcl{D}})
p(f(\mbf{x}^\star) | \mbf{x}^{\star}, \mbf{f}_{\setminus \star}, \mbf{y}_{\mcl{D}})$ is a 2-step process: 
\begin{enumerate}
\item Drawing a sample of $\mbf{f}_{\setminus \star}$ from $p(\mbf{f}_{\setminus \star} | \mbf{y}_{\mcl{D}})$ which is the p.d.f. of a multivariate Gaussian distribution.
\item Given a sample of $\mbf{f}_{\setminus \star}$, drawing a sample of $f(\mbf{x}^\star)$ from $p(f(\mbf{x}^\star)|\mbf{x}^{\star}, \mbf{f}_{\setminus \star}, \mbf{y}_{\mcl{D}})$ which is a lower-truncated Gaussian distribution (truncation of lower tail at $f^+\triangleq\max_{\mbf{x}' \in \mcl{X}^{\star}\setminus\{\mbf{x}^{\star}\}} f(\mbf{x}')$).
\end{enumerate}
The weight of a sample $\mbf{f}_{\mcl{X}^{\star}}$ is $p(\mbf{x}^{\star} | \mbf{f}_{\setminus \star}, \mbf{y}_{\mcl{D}})$ which can be computed efficiently.
\begin{align*}
&p(\mbf{x}^{\star} | \mbf{f}_{\setminus \star}, \mbf{y}_{\mcl{D}}) \\
 &= \int p(\mbf{x}^{\star}, f(\mbf{x}^\star) | \mbf{f}_{\setminus \star}, \mbf{y}_{\mcl{D}})\ \text{d}f(\mbf{x}^\star)\\
 &= \int p(\mbf{x}^{\star}| \mbf{f}_{\mcl{X}^{\star}}, \mbf{y}_{\mcl{D}})
	p(f(\mbf{x}^\star) | \mbf{f}_{\setminus \star}, \mbf{y}_{\mcl{D}})\ \text{d}f(\mbf{x}^\star)\\
 &= \int \mbf{I}_{ f(\mbf{x}^\star) \ge f^+}
	p(f(\mbf{x}^\star) | \mbf{f}_{\setminus \star}, \mbf{y}_{\mcl{D}})\ \text{d}f(\mbf{x}^\star)\\
 &= 1 - \Phi_{ p(f(\mbf{x}^\star) | \mbf{f}_{\setminus \star}, \mbf{y}_{\mcl{D}}) } \left( f^+ \right)
\end{align*}
where $\Phi_{ p(f(\mbf{x}^\star) | \mbf{f}_{\setminus \star}, \mbf{y}_{\mcl{D}}) } \left( f^+ \right)$ is the c.d.f. of the GP predictive belief $p(f(\mbf{x}^\star) | \mbf{f}_{\setminus \star}, \mbf{y}_{\mcl{D}})$ evaluated at $f^+$, and $\mbf{I}_{f(\mbf{x}^\star) \ge f^+}$ is an indicator function such that it is $1$ if $f(\mbf{x}^\star) \ge f^+$ and $0$ otherwise.

\section{Closed-Form Expression of $p(y_{\mbf{x}}| \mbf{f}_{\mcl{X}^{\star}}, \mbf{y}_{\mcl{D}})$}
\label{app:postygvsam}
%

At a BO iteration, let $\mbf{z} \triangleq [\mbf{f}_{\mcl{X}^{\star}}; \mbf{y}_{\mcl{D}}]^\top$ be a column vector where the first $|\mcl{X}^\star|$ elements are $\mbf{f}_{\mcl{X}^{\star}}$ and the last $|\mcl{D}|$ elements are observations $\mbf{y}_{\mcl{D}}$. Let 
$\mbf{t}$ be 
a column vector where the first $|\mcl{X}^\star|$ inputs are $\mcl{X}^{\star}$ and the last $|\mcl{D}|$ inputs are observed inputs $\mcl{D}$. We have the expression: $p(y_{\mbf{x}} | \mbf{f}_{\mcl{X}^{\star}}, \mbf{y}_{\mcl{D}}) = \mcl{N}(y_{\mbf{x}}; \mu_{\mbf{x}}^+, \sigma^2_{\mbf{x}|\mcl{X}^{\star}} + \sigma_n^2)$ specified by
\begin{align}
\mu_{\mbf{x}}^+ 
	&\triangleq \mbf{k}_{\mbf{x} \mbf{t}} 
	\left( 
		\mbf{K}_{\mbf{t} \mbf{t}}
		+ \sigma_n^2 \tilde{\mbf{I}}
	\right)^{-1} 
	\mbf{z}\label{eq:meanplus}\\
\sigma^2_{\mbf{x}|\mcl{X}^{\star}}
	&\triangleq k_{\mbf{x} \mbf{x}}
	- \mbf{k}_{\mbf{x} \mbf{t}} 
	\left( 
		\mbf{K}_{\mbf{t} \mbf{t}}
		+ \sigma_n^2 \tilde{\mbf{I}}
	\right)^{-1}
	\mbf{k}_{\mbf{t} \mbf{x}}\label{eq:varplus}
\end{align}
where $\tilde{\mbf{I}}$ is a matrix of size $(|\mcl{X}^\star| + |\mcl{D}|) \times (|\mcl{X}^\star| + |\mcl{D}|)$ such that $\tilde{\mbf{I}}_{ij} = 1$ if $i = j > |\mcl{X}^\star|$ and $0$ otherwise. Note that $\mu_{\mbf{x}}^+$ and $\sigma_{\mbf{x}|\mcl{X}^{\star}}$ are the mean and standard deviation of $p(f(\mbf{x})| \mbf{f}_{\mcl{X}^\star}, \mbf{y}_{\mcl{D}})$, which are used in Section~\ref{app:eppostpred}.

\section{Expectation Propagation Approximation for $p(\mbf{f}_{\mcl{X}^{\star}} | \mbf{y}_{\mcl{D}}, \mbf{x}^{\star})$}
\label{app:eppostcand}


To approximate $p(\mbf{f}_{\mcl{X}^{\star}} | \mbf{y}_{\mcl{D}}, \mbf{x}^{\star})$, it is expressed as
\begin{align*}
p(\mbf{f}_{\mcl{X}^{\star}} | \mbf{y}_{\mcl{D}}, \mbf{x}^{\star})
	&\propto p(\mbf{f}_{\mcl{X}^{\star}} | \mbf{y}_{\mcl{D}})
	  p(\mbf{x}^{\star}| \mbf{f}_{\mcl{X}^{\star}}, \mbf{y}_{\mcl{D}})\\
	&= p(\mbf{f}_{\mcl{X}^{\star}} | \mbf{y}_{\mcl{D}})
	  \prod_{\mbf{x}' \in \mcl{X}^{\star} \setminus \{\mbf{x}^{\star}\}} \mbf{I}_{f(\mbf{x}^\star) \ge f(\mbf{x}')}\\
	&= p(\mbf{f}_{\mcl{X}^{\star}} | \mbf{y}_{\mcl{D}})
	  \prod_{\mbf{x}' \in \mcl{X}^{\star} \setminus \{\mbf{x}^{\star}\}} \mbf{I}_{\mbf{c}_{\mbf{x}'}^\top \mbf{f}_{\mcl{X}^{\star}} \ge 0}
\end{align*}
where $\mbf{c}_{\mbf{x}}'$ is a column vector of length $|\mcl{X}^\star|$ with $0$ entries except for the $i$-th entry of value $-1$, the $j$-th entry of value $1$ where $i$ and $j$ are the indices of $\mbf{x}'$ and $\mbf{x}^{\star}$ in $\mcl{X}^{\star}$, respectively; $\mbf{I}$ is the indicator function.

The p.d.f. $p(\mbf{f}_{\mcl{X}^{\star}} | \mbf{y}_{\mcl{D}}, \mbf{x}^{\star})$ is approximated with a Gaussian distribution by EP, i.e., $p(\mbf{f}_{\mcl{X}^{\star}} | \mbf{y}_{\mcl{D}}, \mbf{x}^{\star}) \approx \mathcal{N}(\mbf{f}_{\mcl{X}^{\star}}; \bm{\mu}_{\text{ep}}, \bm{\Sigma}_{\text{ep}}) \triangleq q_{\text{ep}}(\mbf{f}_{\mcl{X}^{\star}}|\mbf{y}_{\mcl{D}},\mbf{x}^{\star})$. As in PPES \cite{ppes}, to construct this Gaussian approximation, each indicator term (involved $\mbf{I}$) is approximated with a univariate scaled Gaussian p.d.f.:
\begin{align*}
&q_{\text{ep}}(\mbf{f}_{\mcl{X}^{\star}}|\mbf{y}_{\mcl{D}},\mbf{x}^{\star})
	\triangleq \mathcal{N}(\mbf{f}_{\mcl{X}^{\star}}; \bm{\mu}_{\text{ep}}, \bm{\Sigma}_{\text{ep}})\\
	&\quad= \mathcal{N}(\mbf{f}_{\mcl{X}^{\star}}; \tilde{\mbf{m}}, \tilde{\mbf{K}})
	\prod_{\mbf{x}' \in \mcl{X}^{\star} \setminus \{\mbf{x}^{\star}\}} \tilde{Z}_{\mbf{x}'} \mathcal{N}(\mbf{c}_{\mbf{x}'}^\top \mbf{f}_{\mcl{X}^{\star}}; \tilde{\mu}_{\mbf{x}'}, \tilde{\tau}_{\mbf{x}'})
\end{align*}
where $\mathcal{N}(\mbf{f}_{\mcl{X}^{\star}}; \tilde{\mbf{m}}, \tilde{\mbf{K}})$ denotes the GP predictive belief of $\mbf{f}_{\mcl{X}^{\star}}$ given $\mbf{y}_{\mcl{D}}$, the scale factor $\tilde{Z}_{\mbf{x}'}$ and the variance $\tilde{\tau}_{\mbf{x}'}$ are positive, $\tilde{\mu}_{\mbf{x}'} \in \mbb{R}$. The parameters $\{\tilde{Z}_{\mbf{x}'}, \tilde{\mu}_{\mbf{x}'}, \tilde{\tau}_{\mbf{x}'}\}_{\mbf{x}' \in \mcl{X}^{\star} \setminus \{\mbf{x}^{\star}\}}$ are call the \emph{site parameters}, which are optimized such that the Kullback-Leibler divergence of $q_{\text{ep}}(\mbf{f}_{\mcl{X}^{\star}}|\mbf{y}_{\mcl{D}},\mbf{x}^{\star}) / \int \tilde{p}( \mbf{f}_{\mcl{X}^{\star}})\ \text{d}\mbf{f}_{\mcl{X}^{\star}}$ from $p(\mbf{f}_{\mcl{X}^{\star}} | \mbf{y}_{\mcl{D}}, \mbf{x}^{\star})$ is minimized, i.e., their means and covariance matrices match. As $\{\tilde{Z}_{\mbf{x}'}\}_{\mbf{x}' \in \mcl{X}^{\star} \setminus \{\mbf{x}^{\star}\}}$ are independent from $\mbf{f}_{\mcl{X}^{\star}}$, we only optimize $\{\tilde{\mu}_{\mbf{x}'}, \tilde{\tau}_{\mbf{x}'}\}_{\mbf{x}' \in \mcl{X}^{\star} \setminus \{\mbf{x}^{\star}\}}$.

As the product of Gaussian p.d.f. leads to a Gaussian p.d.f., we have
\begin{align}
\bm{\Sigma}_{\text{ep}} &= \left(
		\tilde{\mbf{K}}^{-1}
		+ \sum_{\mbf{x}' \in \mcl{X}^{\star} \setminus \{\mbf{x}^{\star}\}} \frac{1}{\tilde{\tau}_{\mbf{x}'}} \mbf{c}_{\mbf{x}'} \mbf{c}_{\mbf{x}'}^\top	
	\right)^{-1}\label{eq:epmean}\\
\bm{\mu}_{\text{ep}} &= \bm{\Sigma}_{\text{ep}} \left(
		\tilde{\mbf{K}}^{-1} \tilde{\mbf{m}}
		+ \sum_{\mbf{x}' \in \mcl{X}^{\star} \setminus \{\mbf{x}^{\star}\}} \frac{\tilde{\mu}_{\mbf{x}'}}{\tilde{\tau}_{\mbf{x}'}} \mbf{c}_{\mbf{x}'}	
	\right)\ .\label{eq:epcov}
\end{align}
To update the site parameters, we first compute the \emph{cavity distributions}
\begin{align*}
p_{\setminus \mbf{x}'}(\mbf{f}_{\mcl{X}^{\star}}) \triangleq \frac{q_{\text{ep}}(\mbf{f}_{\mcl{X}^{\star}}|\mbf{y}_{\mcl{D}},\mbf{x}^{\star})}{\tilde{Z}_{\mbf{x}'} \mathcal{N}(\mbf{c}_{\mbf{x}'}^\top \mbf{f}_{\mcl{X}^{\star}}; \tilde{\mu}_{\mbf{x}'}, \tilde{\tau}_{\mbf{x}'})}\ .
\end{align*}
Note that in the next step, we would like to approximate the distribution of $\mbf{c}_{\mbf{x}'}^\top \mbf{f}_{\mcl{X}^\star}$, so we only require the cavity distribution of the r.v. $\mbf{c}_{\mbf{x}'}^\top \mbf{f}_{\mcl{X}^\star}$ which is denoted as:
\begin{align*}
p_{\setminus {\mbf{x}'}}(\mbf{c}_{\mbf{x}'}^\top \mbf{f}_{\mcl{X}^{\star}}) \triangleq \frac{\tilde{p}(\mbf{c}_{\mbf{x}'}^\top \mbf{f}_{\mcl{X}^{\star}})}{\tilde{Z}_{\mbf{x}'} \mathcal{N}(\mbf{c}_{\mbf{x}'}^\top \mbf{f}_{\mcl{X}^{\star}}; \tilde{\mu}_{\mbf{x}'}, \tilde{\tau}_{\mbf{x}'})}
\end{align*}
where $\tilde{p}(\mbf{c}_{\mbf{x}'}^\top \mbf{f}_{\mcl{X}^{\star}}) = \mathcal{N}(\mbf{c}_{\mbf{x}'}^\top \mbf{f}_{\mcl{X}^{\star}}; \mbf{c}_{\mbf{x}'}^\top \bm{\mu}_{\text{ep}}, \mbf{c}_{\mbf{x}'}^\top \bm{\Sigma}_{\text{ep}} \mbf{c}_{\mbf{x}'})$. Let the Gaussian mean and variance of $p_{\setminus {\mbf{x}'}}(\mbf{c}_{\mbf{x}'}^\top \mbf{f}_{\mcl{X}^{\star}})$ are $\mu_{\setminus \mbf{x}'}$ and $\tau_{\setminus \mbf{x}'}$, respectively, we have
\begin{align*}
\tau_{\setminus \mbf{x}'} &= \left( \left(\mbf{c}_{\mbf{x}'}^\top \bm{\Sigma}_{\text{ep}} \mbf{c}_{\mbf{x}'}\right)^{-1} - \tilde{\tau}_{\mbf{x}'}^{-1} \right)^{-1}\\
\mu_{\setminus \mbf{x}'} &= \tau_{\setminus \mbf{x}'} \left( 
	\frac{\mbf{c}_{\mbf{x}'}^\top \bm{\mu}_{\text{ep}}}{\mbf{c}_{\mbf{x}'}^\top \bm{\Sigma}_{\text{ep}} \mbf{c}_{\mbf{x}'}} 
	- \frac{\tilde{\mu}_{\mbf{x}'}}{\tilde{\tau}_{\mbf{x}'}} \right)\ .
\end{align*}
Next, the \emph{projection step} of EP is to do moment matching of $\tilde{Z}_{\mbf{x}'} \mathcal{N}(\mbf{c}_{\mbf{x}'}^\top \mbf{f}_{\mcl{X}^{\star}}; \tilde{\mu}_{\mbf{x}'}, \tilde{\tau}_{\mbf{x}'}) p_{\setminus \mbf{x}'}(\mbf{c}_{\mbf{x}'}^\top \mbf{f}_{\mcl{X}^{\star}})$ with $\mbf{I}_{\mbf{c}_{\mbf{x}'}^\top \mbf{f}_{\mcl{X}^{\star}} \ge 0}\ p_{\setminus \mbf{x}'}(\mbf{c}_{\mbf{x}'}^\top \mbf{f}_{\mcl{X}^{\star}})$. We use the derivatives of the zeroth moment to compute moments of $\mbf{I}_{\mbf{c}_{\mbf{x}'}^\top \mbf{f}_{\mcl{X}^{\star}} \ge 0}\ p_{\setminus \mbf{x}'}(\mbf{c}_{\mbf{x}'}^\top \mbf{f}_{\mcl{X}^{\star}})$, denoted as $\hat{\mu}_\mbf{x}$ (mean) and $\hat{\tau}_\mbf{x}$ (variance). The zeroth moment is computed as:
\begin{align}
\hat{Z}_{\mbf{x}'} 
	&\triangleq \int \mbf{I}_{\mbf{c}_{\mbf{x}'}^\top \mbf{f}_{\mcl{X}^{\star}} \ge 0}\ p_{\setminus \mbf{x}'}(\mbf{c}_{\mbf{x}'}^\top \mbf{f}_{\mcl{X}^{\star}})\ \text{d}\left(\mbf{c}_{\mbf{x}'}^\top \mbf{f}_{\mcl{X}^{\star}}\right)\nonumber\\
	&= \int \mbf{I}_{\mbf{c}_{\mbf{x}'}^\top \mbf{f}_{\mcl{X}^{\star}} \ge 0}\ \mathcal{N}(\mbf{c}_{\mbf{x}'}^\top \mbf{f}_{\mcl{X}^{\star}}; \mu_{\setminus \mbf{x}'}, \tau_{\setminus \mbf{x}'})\ \text{d}\left(\mbf{c}_{\mbf{x}'}^\top \mbf{f}_{\mcl{X}^{\star}}\right)\nonumber\\
	&= \Phi(\beta_{\mbf{x}'})\ \text{where}\ \beta_{\mbf{x}'} \triangleq \frac{\mu_{\setminus \mbf{x}'}}{\sqrt{\tau_{\setminus \mbf{x}'}}}\nonumber\\
\frac{\partial \hat{Z}_{\mbf{x}'}}{\partial \mu_{\setminus \mbf{x}'}}
	&= \frac{\Phi(\beta_{\mbf{x}'})}{\partial \mu_{\setminus \mbf{x}'}} 
	= \frac{\phi(\beta_{\mbf{x}'})}{\sqrt{\tau_{\setminus \mbf{x}'}}}\label{eq:dmzero1}\\
\frac{\partial \hat{Z}_{\mbf{x}'}}{\partial \tau_{\setminus \mbf{x}'}}
	&= \frac{\Phi(\beta_{\mbf{x}'})}{\partial \tau_{\setminus \mbf{x}'}}
	= -\frac{1}{2} \frac{\mu_{\setminus \mbf{x}'}}{\sqrt{\tau_{\setminus \mbf{x}'}^3}} \phi(\beta_{\mbf{x}'})\label{eq:dvzero1}
\end{align}
where $\phi$ and $\Phi$ are the p.d.f. and c.d.f. of the standard Gaussian distribution, respectively. On the other hand, we can express the derivatives of $\hat{Z}_{\mbf{x}'}$ as:
\begin{align}
&\frac{\partial \hat{Z}_{\mbf{x}'}}{\partial \mu_{\setminus \mbf{x}'}}\\
	&\quad= \frac{\partial}{\partial \mu_{\setminus \mbf{x}'}} \int \mbf{I}_{\mbf{c}_{\mbf{x}'}^\top \mbf{f}_{\mcl{X}^{\star}} \ge 0}\ \mathcal{N}(\mbf{c}_{\mbf{x}'}^\top \mbf{f}_{\mcl{X}^{\star}}; \mu_{\setminus \mbf{x}'}, \tau_{\setminus \mbf{x}'})\ \text{d}\left(\mbf{c}_{\mbf{x}'}^\top \mbf{f}_{\mcl{X}^{\star}}\right)\nonumber\\
	&\quad= \hat{Z}_{\mbf{x}'} \frac{\hat{\mu}_{\mbf{x}'}}{\tau_{\setminus \mbf{x}'}}
	- \hat{Z}_{\mbf{x}'} \frac{\mu_{\setminus \mbf{x}'}}{\tau_{\setminus \mbf{x}'}}\nonumber\\
	&\quad= \Phi(\beta_{\mbf{x}'})\frac{\hat{\mu}_{\mbf{x}'}}{\tau_{\setminus \mbf{x}'}}
	- \Phi(\beta_{\mbf{x}'})\frac{\mu_{\setminus \mbf{x}'}}{\tau_{\setminus \mbf{x}'}}\label{eq:dmzero2}\\
&\frac{\partial \hat{Z}_{\mbf{x}'}}{\partial \tau_{\setminus \mbf{x}'}}\\
	&\quad= \frac{\partial}{\partial \tau_{\setminus \mbf{x}'}} \int \mbf{I}_{\mbf{c}_{\mbf{x}'}^\top \mbf{f}_{\mcl{X}^{\star}} \ge 0}\ \mathcal{N}(\mbf{c}_{\mbf{x}'}^\top \mbf{f}_{\mcl{X}^{\star}}; \mu_{\setminus \mbf{x}'}, \tau_{\setminus \mbf{x}'})\ \text{d}\left(\mbf{c}_{\mbf{x}'}^\top \mbf{f}_{\mcl{X}^{\star}}\right)\nonumber\\
	&\quad= \frac{1}{2} \hat{Z}_{\mbf{x}'} \frac{\hat{\mu}_{\mbf{x}'}^2 + \hat{\tau}_{\mbf{x}'}}{\tau_{\setminus \mbf{x}'}^2}
	- \hat{Z}_{\mbf{x}'} \frac{\mu_{\setminus \mbf{x}'}}{\tau_{\setminus \mbf{x}'}^2} \hat{\mu}_{\mbf{x}'}
	+ \frac{1}{2} \hat{Z}_{\mbf{x}'} \frac{\mu_{\setminus \mbf{x}'}^2}{\tau_{\setminus \mbf{x}'}^2} 
	- \frac{1}{2} \frac{\hat{Z}_{\mbf{x}'} }{\tau_{\setminus \mbf{x}'}}\nonumber\\
	&\quad= \frac{1}{2} \Phi(\beta_{\mbf{x}'}) \frac{\hat{\mu}_{\mbf{x}'}^2 + \hat{\tau}_{\mbf{x}'}}{\tau_{\setminus \mbf{x}'}^2}
	- \Phi(\beta_{\mbf{x}'}) \frac{\mu_{\setminus \mbf{x}'}}{\tau_{\setminus \mbf{x}'}^2} \hat{\mu}_{\mbf{x}'}\nonumber\\
	&\quad\quad+ \frac{1}{2} \Phi(\beta_{\mbf{x}'}) \frac{\mu_{\setminus \mbf{x}'}^2}{\tau_{\setminus \mbf{x}'}^2} 
	- \frac{1}{2} \frac{\Phi(\beta_{\mbf{x}'})}{\tau_{\setminus \mbf{x}'}}\ .\label{eq:dvzero2}
\end{align}
Equating~\eqref{eq:dmzero1} with~\eqref{eq:dmzero2}, and~\eqref{eq:dvzero1} with~\eqref{eq:dvzero2}, we have
\begin{align*}
\hat{\mu}_{\mbf{x}'} &= \sqrt{\tau_{\setminus \mbf{x}'}} \frac{\phi(\beta_{\mbf{x}'})}{\Phi(\beta_{\mbf{x}'})} + \mu_{\setminus \mbf{x}'}\\
\hat{\tau}_{\mbf{x}'} &= -\mu_{\setminus \mbf{x}'} \sqrt{\tau_{\setminus \mbf{x}'}} \frac{\phi(\beta_{\mbf{x}'})}{\Phi(\beta_{\mbf{x}'})}
	- \hat{\mu}_{\mbf{x}'}^2 
	+ 2\mu_{\setminus \mbf{x}'} \hat{\mu}_{\mbf{x}'}
	- \mu_{\setminus \mbf{x}'}^2
	+ \tau_{\setminus \mbf{x}'}\ .
\end{align*}
Then, we update the site parameters to get the moments of $\tilde{Z}_{\mbf{x}'} \mathcal{N}(\mbf{c}_{\mbf{x}'}^\top \mbf{f}_{\mcl{X}^{\star}}; \tilde{\mu}_{\mbf{x}'}, \tilde{\tau}_{\mbf{x}'}) = \mbf{I}_{\mbf{c}_{\mbf{x}'}^\top \mbf{f}_{\mcl{X}^{\star}} \ge 0}\ p_{\setminus \mbf{x}'}(\mbf{c}_{\mbf{x}'}^\top \mbf{f}_{\mcl{X}^{\star}}) / p_{\setminus \mbf{x}'}(\mbf{c}_{\mbf{x}'}^\top \mbf{f}_{\mcl{X}^{\star}})$ as
\begin{align*}
\tilde{\tau}_{\mbf{x}'} &= \left(\hat{\tau}_{\mbf{x}'}^{-1} - \tau_{\setminus \mbf{x}'}^{-1}\right)^{-1}\\
\tilde{\mu}_{\mbf{x}'} &= \tilde{\tau}_{\mbf{x}'} \left(
	\hat{\tau}_{\mbf{x}'}^{-1} \hat{\mu}_{\mbf{x}'}
	- \tau_{\setminus \mbf{x}'}^{-1} \mu_{\setminus \mbf{x}'}
\right)\ .
\end{align*}
Finally, we update the parameters $\bm{\mu}_{\text{ep}}$ and $\bm{\Sigma}_{\text{ep}}$ by~\eqref{eq:epmean} and~\eqref{eq:epcov}. The process is repeated for all $\mbf{x}' \in \mcl{X}^{\star} \setminus \{\mbf{x}^{\star}\}$ until convergence.

\section{Closed-Form Expression of $p(f(\mbf{x}) | \mbf{y}_{\mcl{D}}, \mbf{x}^{\star})$}
\label{app:eppostpred}

We can express the predictive posterior distribution $p(f(\mbf{x}) | \mbf{y}_{\mcl{D}}, \mbf{x}^{\star})$ as:
\begin{align*}
&p(f(\mbf{x}) | \mbf{y}_{\mcl{D}}, \mbf{x}^{\star})\\
	&= \int p(f(\mbf{x})| \mbf{f}_{\mcl{X}^{\star}}, \mbf{y}_{\mcl{D}}, \mbf{x}^{\star})
		p(\mbf{f}_{\mcl{X}^{\star}}| \mbf{y}_{\mcl{D}}, \mbf{x}^{\star})\ \text{d}\mbf{f}_{\mcl{X}^{\star}}\\
	&= \int p(f(\mbf{x})| \mbf{f}_{\mcl{X}^{\star}}, \mbf{y}_{\mcl{D}})
		p(\mbf{f}_{\mcl{X}^{\star}}| \mbf{y}_{\mcl{D}}, \mbf{x}^{\star})\ \text{d}\mbf{f}_{\mcl{X}^{\star}}\\
	&\approx \int p(f(\mbf{x})| \mbf{f}_{\mcl{X}^{\star}}, \mbf{y}_{\mcl{D}})
		q_{\text{ep}}(\mbf{f}_{\mcl{X}^{\star}}|\mbf{y}_{\mcl{D}},\mbf{x}^{\star})\ \text{d}\mbf{f}_{\mcl{X}^{\star}}\\
	&= \int \mathcal{N}(f(\mbf{x}); \mu_{\mbf{x}}^+, \sigma^2_{\mbf{x}|\mcl{X}^{\star}}) 
		\mathcal{N}(\mbf{f}_{\mcl{X}^{\star}}; \bm{\mu}_{\text{ep}}, \bm{\Sigma}_{\text{ep}})\ \text{d}\mbf{f}_{\mcl{X}^{\star}}\ .
\end{align*}
where $p(f(\mbf{x})| \mbf{f}_{\mcl{X}^{\star}}, \mbf{y}_{\mcl{D}}) \triangleq \mathcal{N}(f(\mbf{x}); \mu_{\mbf{x}}^+, \sigma^2_{\mbf{x}|\mcl{X}^{\star}})$; $\mu_{\mbf{x}}^+$ and $\sigma^2_{\mbf{x}|\mcl{X}^{\star}}$ are defined in Eq.~\ref{eq:meanplus} and Eq.~\ref{eq:varplus}, respectively.


Let $\mbf{r} \triangleq \mbf{K}_+^{-1} \mbf{k}_{\mbf{t} \mbf{x}}$, $\mbf{a} \triangleq \mbf{r}_{\mcl{X}^\star}$, and $b \triangleq \mbf{r}_{\mcl{D}}^\top \mbf{y}_{\mcl{D}}$ where $\mbf{r}_{\mcl{X}^\star}$ and $\mbf{r}_{\mcl{D}}$ are the first $|\mcl{X}^\star|$ and the last $|\mcl{D}|$ elements of the column vector $\mbf{r}$.
Then, we have
\begin{align*}
\mu_{\mbf{x}}^+ &= \mbf{a}^\top \mbf{f}_{\mcl{X}^{\star}} + b\\
p(f(\mbf{x}) | \mbf{y}_{\mcl{D}}, \mbf{x}^{\star}) &\approx \int \mathcal{N}(f(\mbf{x}); \mbf{a}^\top \mbf{f}_{\mcl{X}^{\star}} + b, \sigma^2_{\mbf{x}|\mcl{X}^{\star}}) \\
	&\quad\times \mathcal{N}(\mbf{f}_{\mcl{X}^{\star}}; \bm{\mu}_{\text{ep}}, \bm{\Sigma}_{\text{ep}})\ \text{d}\mbf{f}_{\mcl{X}^{\star}}\\
	&= \mcl{N}(f(\mbf{x}); \mbf{a}^\top \bm{\mu}_{\text{ep}} + b, 
		\sigma^2_{\mbf{x}|\mcl{X}^{\star}} + \mbf{a}^\top \bm{\Sigma}_{\text{ep}} \mbf{a})\ .
\end{align*}
Hence, 
\begin{align*}
p(y_{\mbf{x}} | \mbf{y}_{\mcl{D}}, \mbf{x}^{\star})
	&\approx \mcl{N}(f(\mbf{x}); \mbf{a}^\top \bm{\mu}_{\text{ep}} + b, 
		\sigma^2_{\mbf{x}|\mcl{X}^{\star}} + \mbf{a}^\top \bm{\Sigma}_{\text{ep}} \mbf{a} + \sigma_n^2)\\
	&\triangleq q_{\text{ep}}(y_{\mbf{x}}|\mbf{y}_{\mcl{D}}, \mbf{x}^{\star})\ .
\end{align*}

\section{An Example on Exploitation vs. Exploration of Different TES Approximation Methods}
\label{app:epexploit}

\begin{figure}
\hspace{-5.5mm}
\centering
\includegraphics[height=0.27\textwidth]{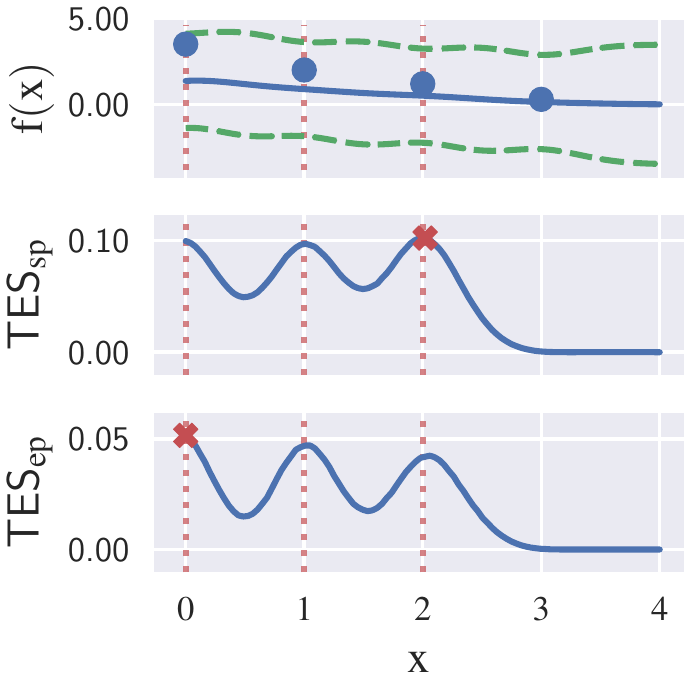}
\caption{An example: TES\textsubscript{ep} exploits more than TES\textsubscript{sp}. The top plot shows the GP posterior mean as a solid blue line, uncertainty (variance) as dashed green lines, and data points as blue points. The dotted red lines show the positions of $\mcl{X}^{\star}$. The red crosses indicate the maximizers of acquisition functions.}
\label{fig:epexploit}
\end{figure}

Fig.~\ref{fig:epexploit} shows an example where TES\textsubscript{sp} and TES\textsubscript{ep} select different inputs. In particular, TES\textsubscript{ep} selects an input where the GP posterior mean of its function value is higher than that of the input TES\textsubscript{sp} selects. Scrutinizing more carefully on the acquisition function values $f(0)$ and $f(2)$, we observe that the difference in values of TES\textsubscript{sp} is smaller than that of TES\textsubscript{ep}. 
In these cases, it means that TES\textsubscript{ep} has a stronger preference to a larger mean function value than TES\textsubscript{sp}. 
Hence, it is observed in these cases that TES\textsubscript{ep} exploits more than TES\textsubscript{sp}.

\section{TES\textsubscript{ep} of Different Batch Sizes}
\label{app:diffbatchsize}

\begin{figure}
\centering
\includegraphics[height=0.18\textwidth]{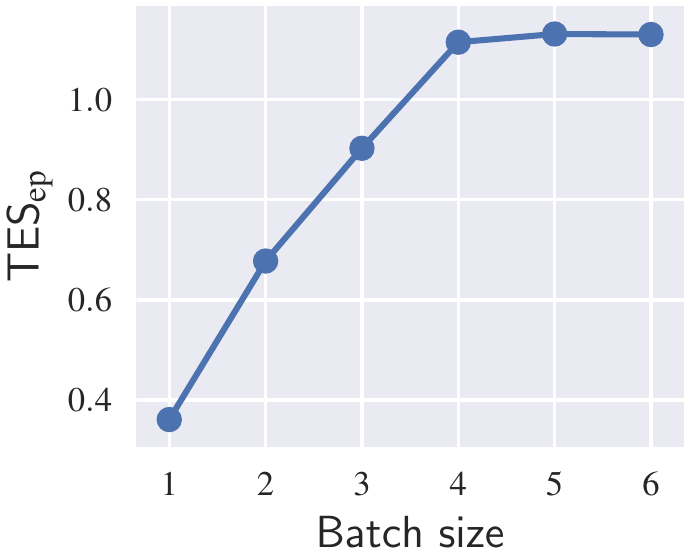}
\caption{Plot of TES\textsubscript{ep} against the batch size for $n = 5$.}
\label{fig:batchsizenmax}
\end{figure}

Note that $\alpha^{\star}(\mbf{y}_{\mcl{D}}, \mcl{B})$ is the information gain about $\mbf{x}^{\star}$ through observing $\mbf{y}_{\mcl{B}}$. Thus, it is comparable between batches of different sizes. Given the size of $\mcl{X}^\star$ as $5$, Fig.~\ref{fig:batchsizenmax} shows the maximum TES\textsubscript{ep} value (i.e., maximum information gain) for different batch sizes. We observe that increasing the batch size to be larger than $|\mcl{X}^\star|$ only yields an insignificant amount of TES\textsubscript{ep}. Therefore, the size of $\mcl{X}^\star$ should be at least the batch size to make the most out of the observations $\mbf{y}_{\mcl{B}}$. Besides, we can select the batch size adaptively at each BO iteration based on a trade-off between the increase in the information gain and the cost of an observation.

\section{Additional Experiment Results}
\label{app:experiment}

\begin{figure}
\hspace{-2mm}
\begin{tabular}{cc}	
	\includegraphics[height=0.21\textwidth]{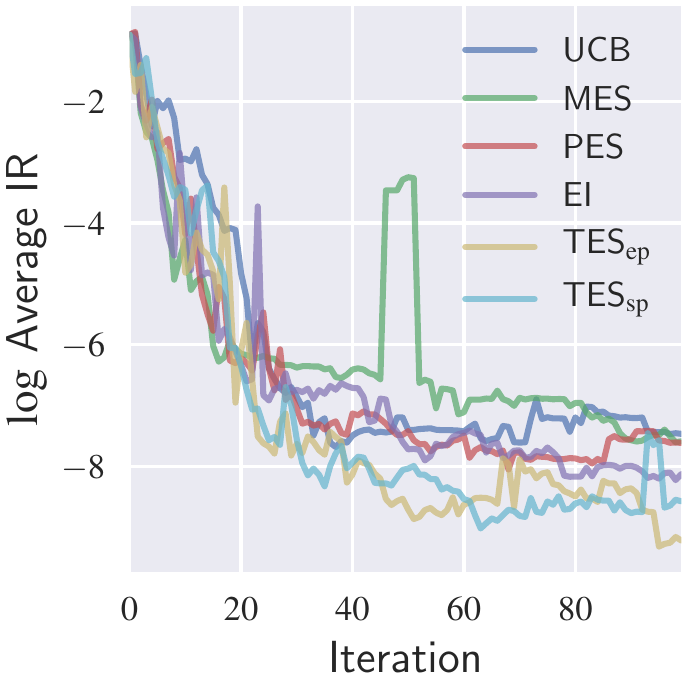}
	&
	\includegraphics[height=0.21\textwidth]{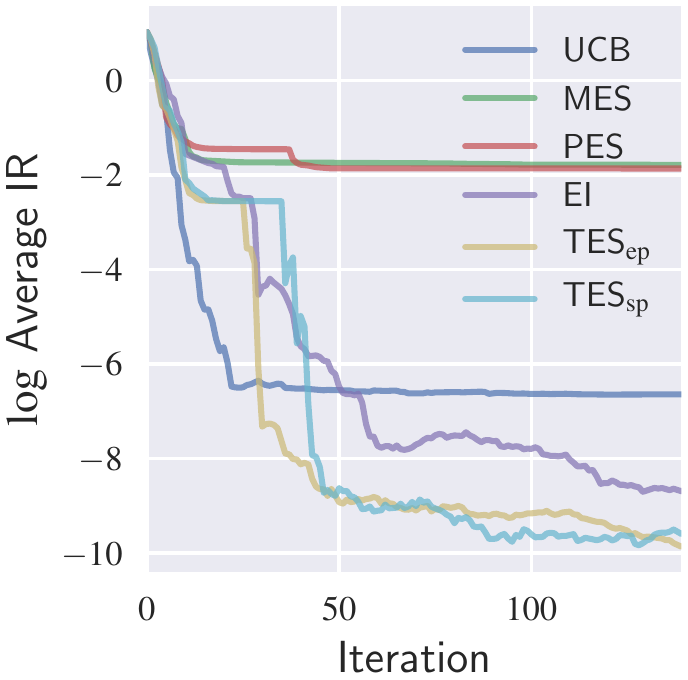}
	\\
	(a) Branin.
	&
	(b) Hartmann-3D.
\end{tabular}
\caption{Additional BO results.}
\label{fig:expsync}
\end{figure}

Fig.~\ref{fig:expsync} shows the average of the results over $10$ random runs for $2$ synthetic functions: Branin defined in $[0,1]^2$, and Hartmann-3D defined in $[0,1]^3$ \citep{lizotte08}.
As our optimization problem is maximization and these synthetic functions are often minimized, we take the negative values of these functions as objective functions. 
In these experiments, the noise variance is fixed to $\sigma_n^2 = 10^{-4}$. 
Note that unlike the synthetic function we sample from a GP posterior, the assumption of using an \emph{isotropic} SE kernel\footnote{A kernel $k_{\mbf{x}\mbf{x}'}$ is \emph{isotropic} if it depends on $|\mbf{x} - \mbf{x}'|$ only.} might be violated in these synthetic functions. However, the empirical results in Fig.~\ref{fig:expsync} still show a reasonable performance. Overall, TES acquisition functions outperform others, but the difference is less obvious in the Branin function. It could be because Branin is a simple function to optimize (i.e., having $3$ local maxima in a 2-dimensional input space). In the Hartmann-3D experiment, EI outperforms PES, which could be because PES explores more than EI as explained by \citet{pes}.

\begin{figure}[ht]
\hspace{-2mm}
\begin{tabular}{cc}	
	\includegraphics[height=0.2\textwidth]{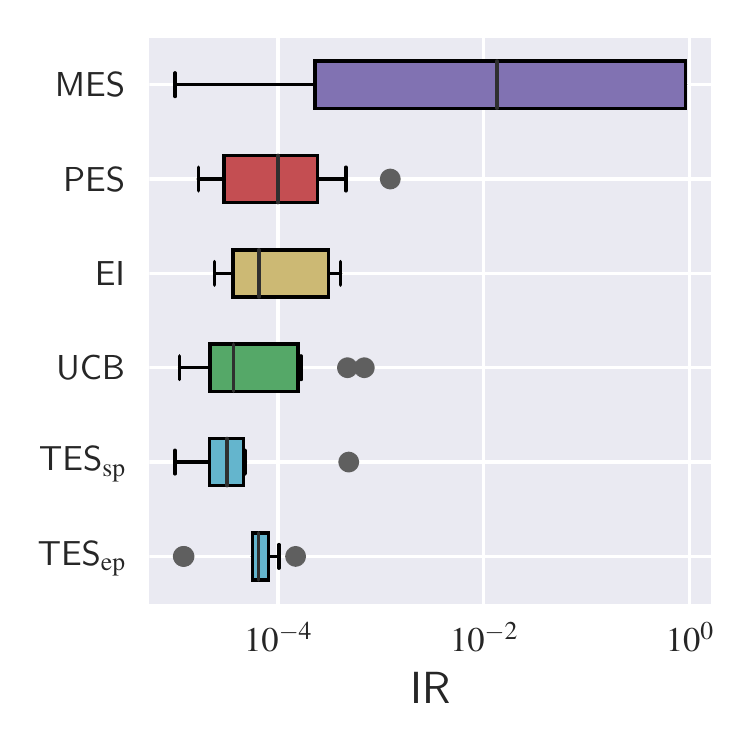}
	&
	\includegraphics[height=0.2\textwidth]{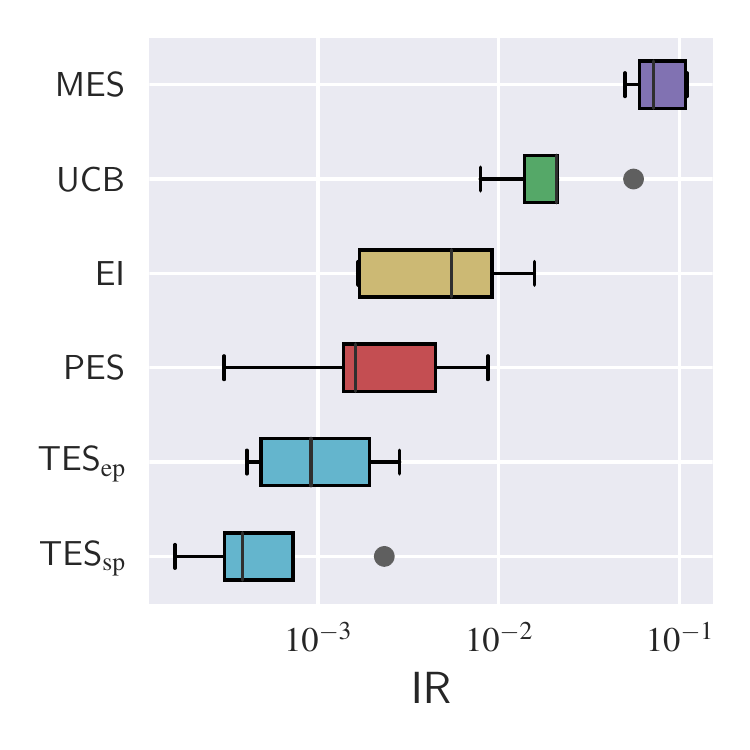}
	\\
	(a) GP sampled function.
	&
	(b) log10P.
	\\
	\includegraphics[height=0.2\textwidth]{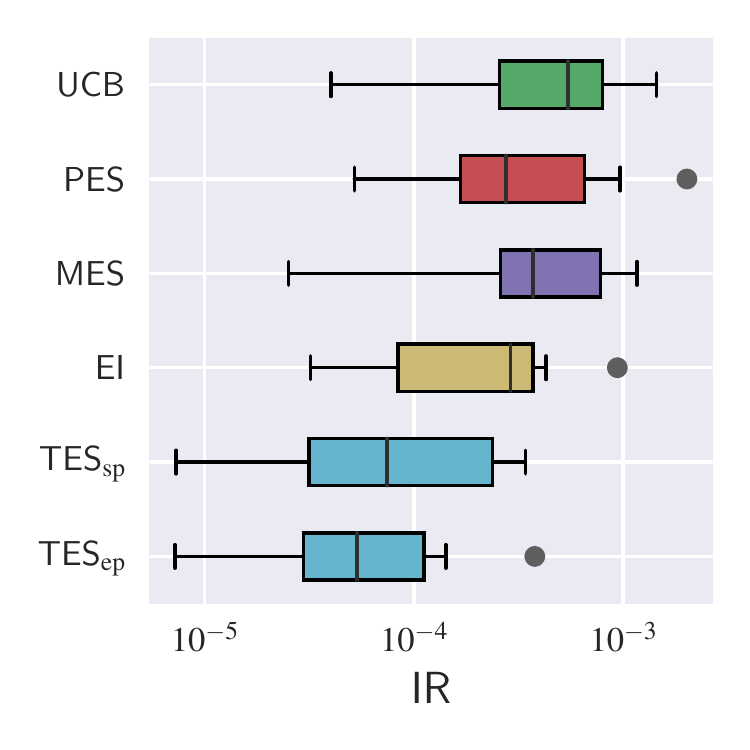}
	&
	\includegraphics[height=0.2\textwidth]{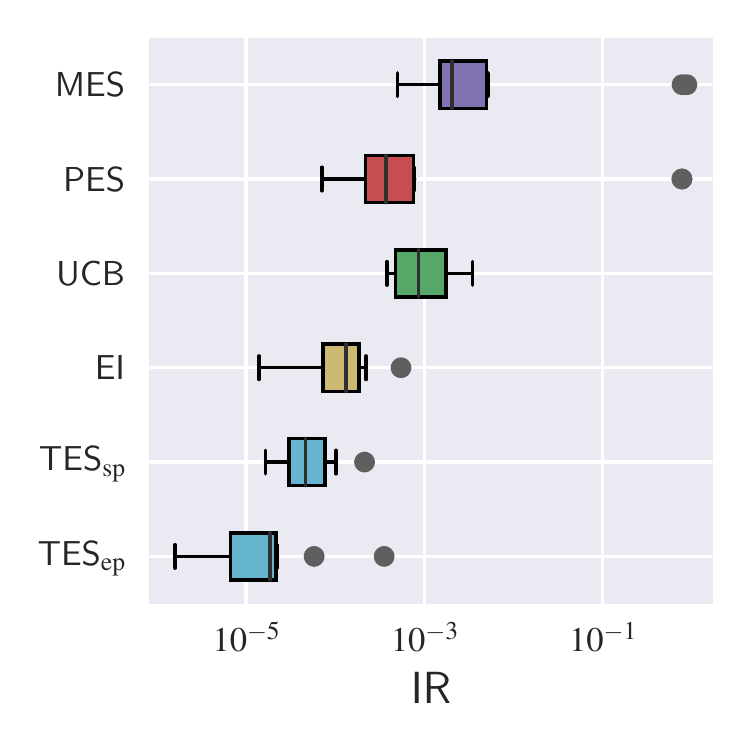}
	\\
	(c) Branin.
	&
	(d) Hartmann-3D.
\end{tabular}
\caption{Box plots of BO with $|\mcl{B}| = 1$.}
\label{fig:expsyncboxplot}
\end{figure}

\begin{figure}[ht!]
\begin{tabular}{cc}	
	\includegraphics[height=0.2\textwidth]{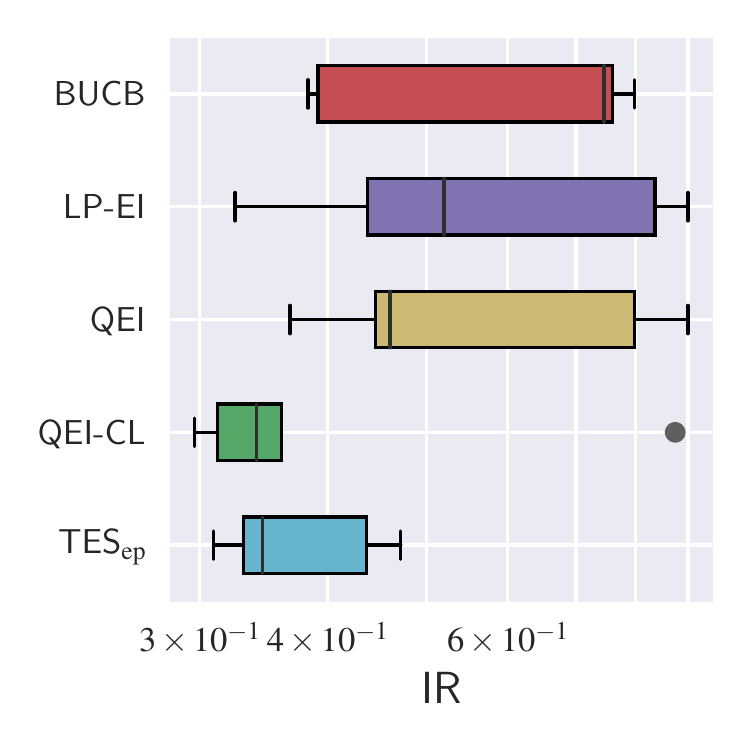}
	&
	\includegraphics[height=0.2\textwidth]{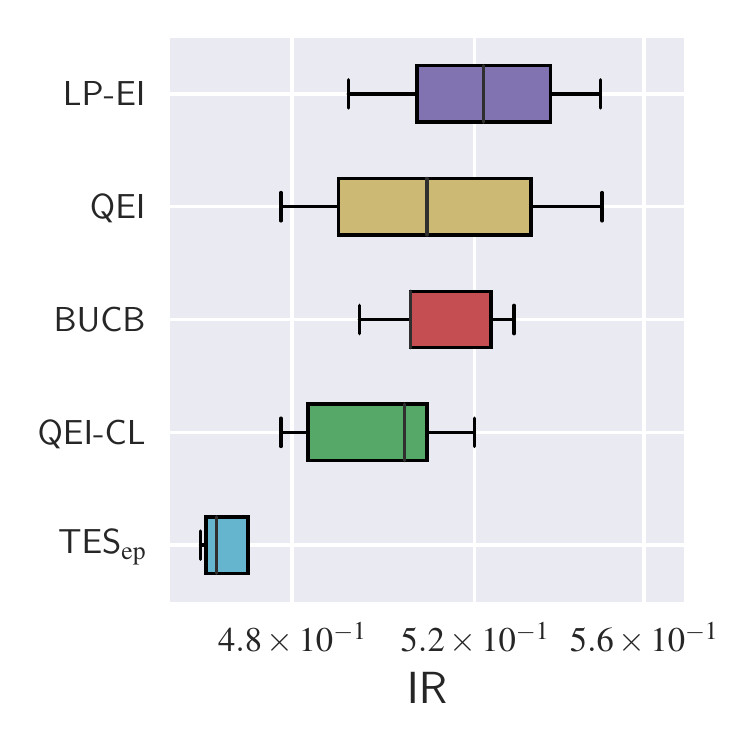}
	\\
	(a) The CIFAR-10.
	&
	(b) The face attack function.
\end{tabular}
\caption{Box plots of batch BO results for CIFAR-10 and the face attack function. $|\mcl{B}| = 20$.}
\label{fig:realworldboxplot}
\end{figure}

\begin{figure}[ht]
\hspace{-2mm}
\begin{tabular}{@{}c@{}c@{}}	
	\includegraphics[height=0.2\textwidth]{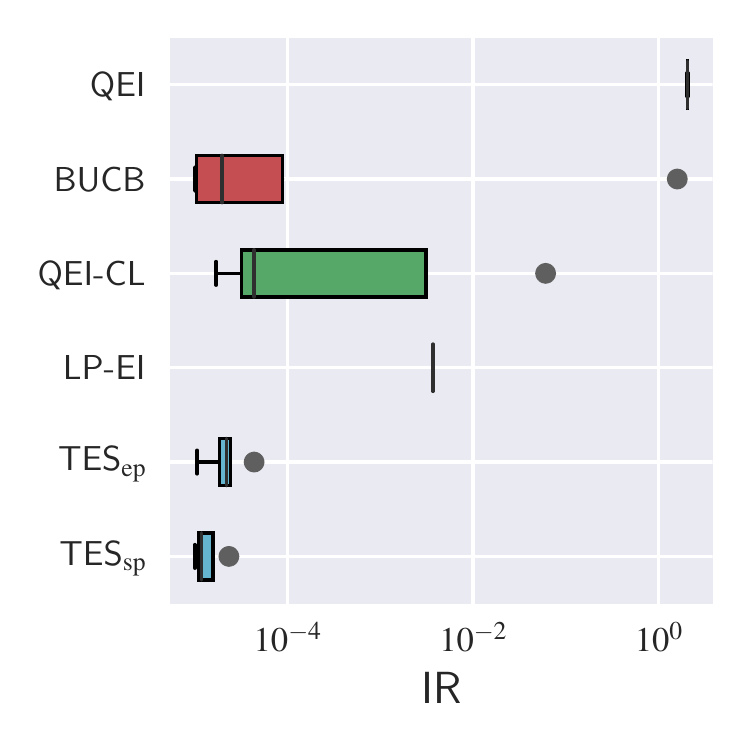}
	&
	\includegraphics[height=0.2\textwidth]{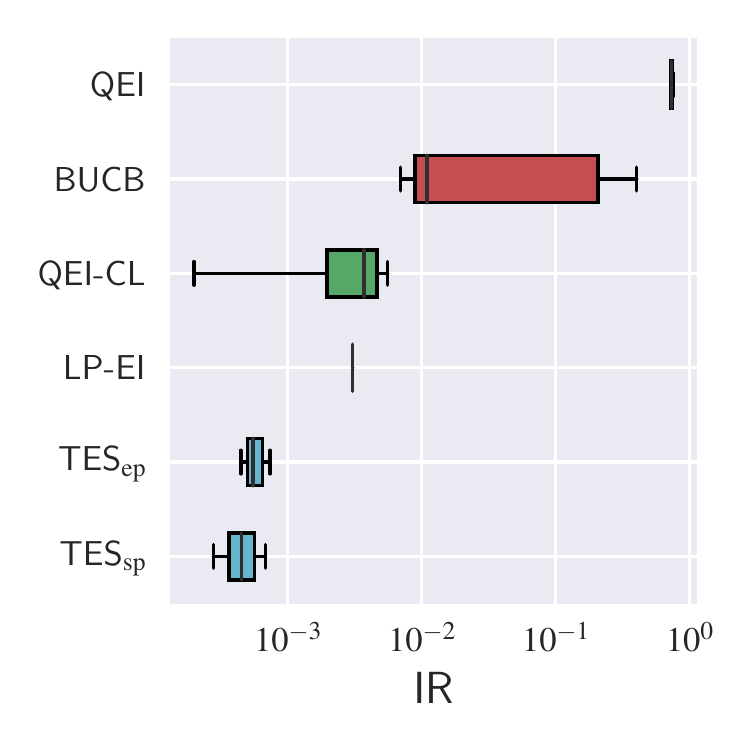}
	\\
	(a) GP sampled function $|\mcl{B}| = 3$.
	&
	(b) Hartmann-4D $|\mcl{B}| = 3$.
	\\
	\includegraphics[height=0.2\textwidth]{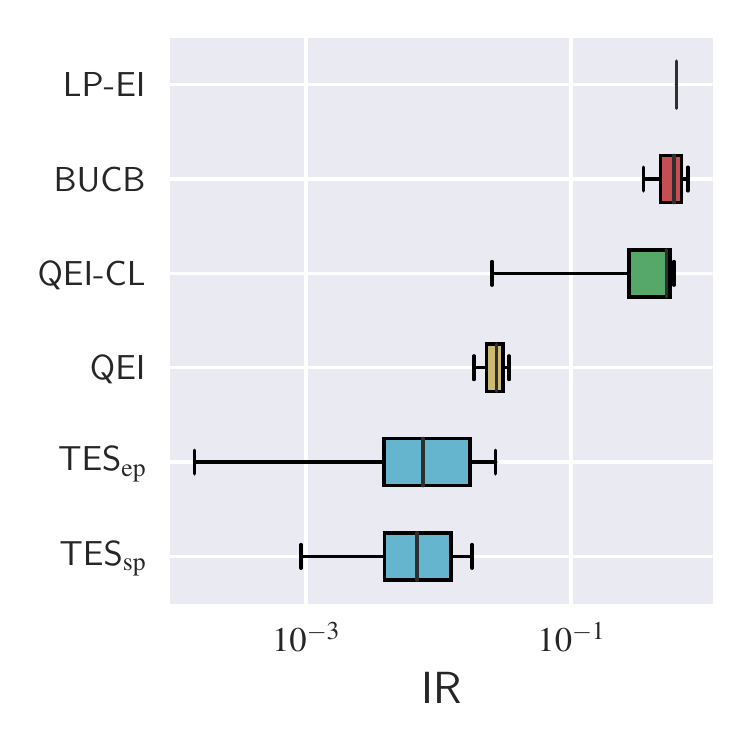}
	&
	\includegraphics[height=0.2\textwidth]{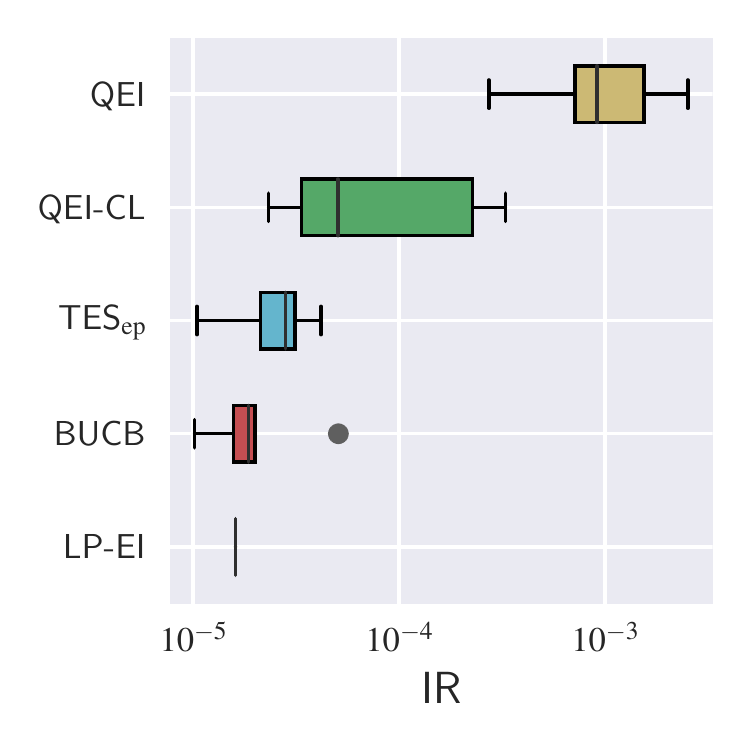}
	\\
	(c) log10P $|\mcl{B}| = 3$.
	&
	(d) GP sampled function \\
	& $|\mcl{B}| = 10$.
	\\
	\includegraphics[height=0.2\textwidth]{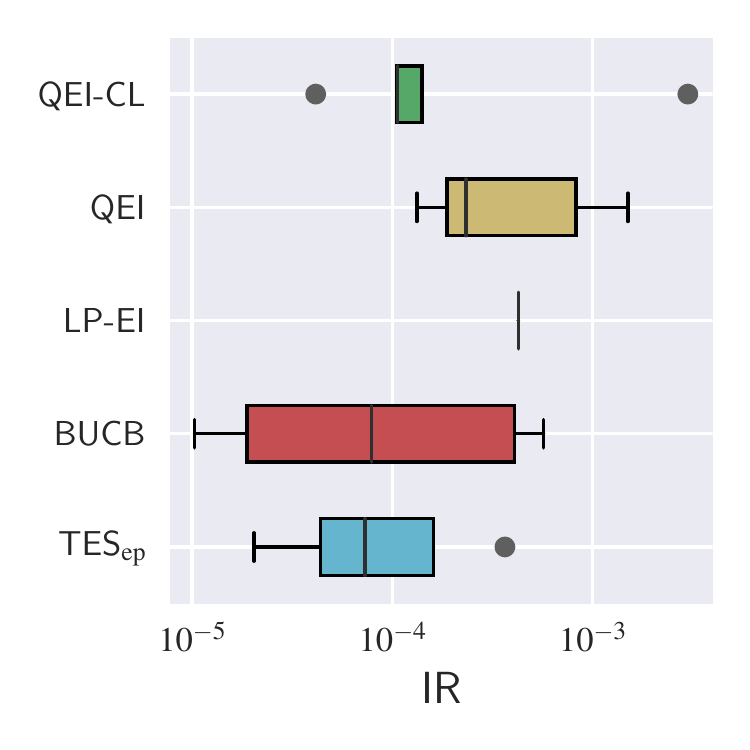}
	&
	\includegraphics[height=0.2\textwidth]{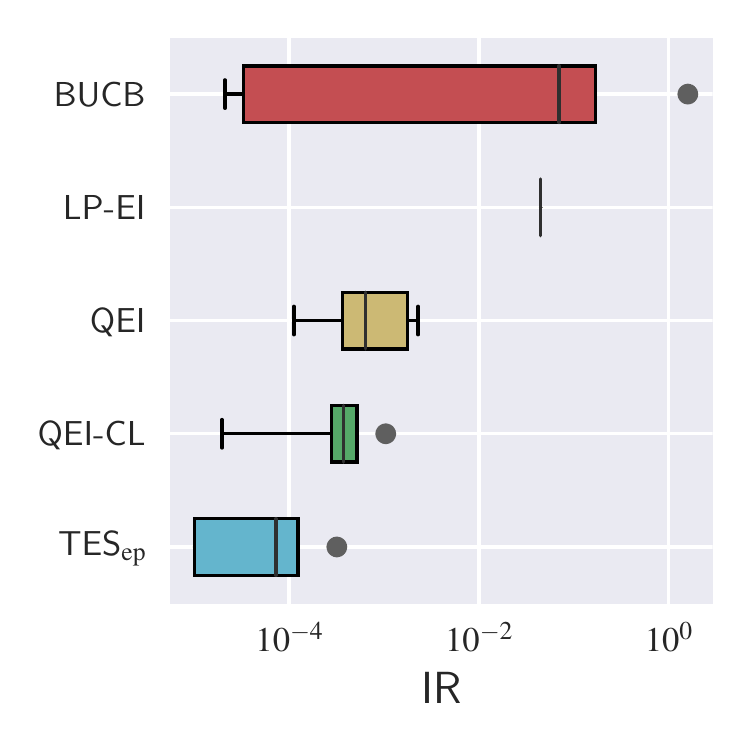}
	\\
	(e) GP sampled function & (f) GP sampled function\\
	$|\mcl{B}| = 20$. & $|\mcl{B}| = 30$.
	\\
	\includegraphics[height=0.2\textwidth]{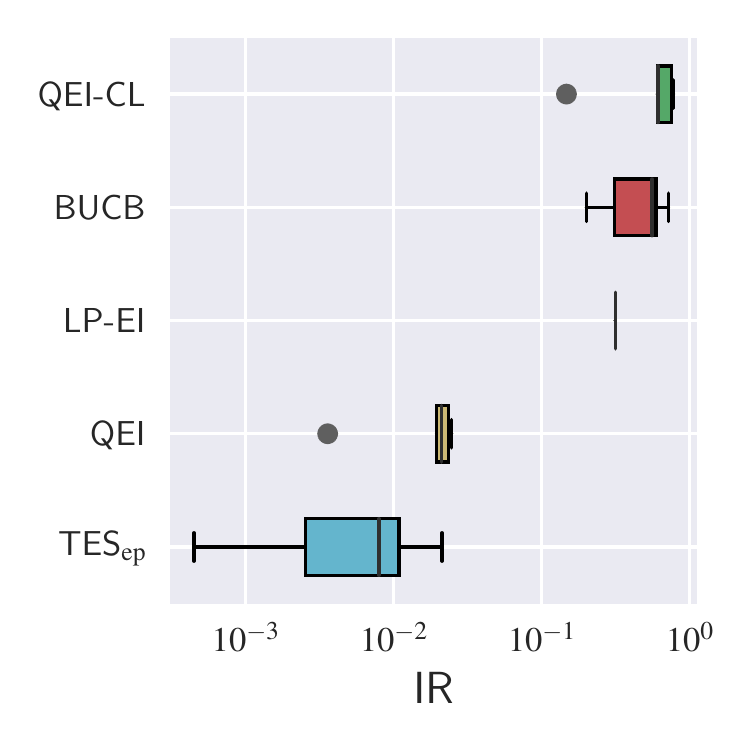}
	&
	\includegraphics[height=0.2\textwidth]{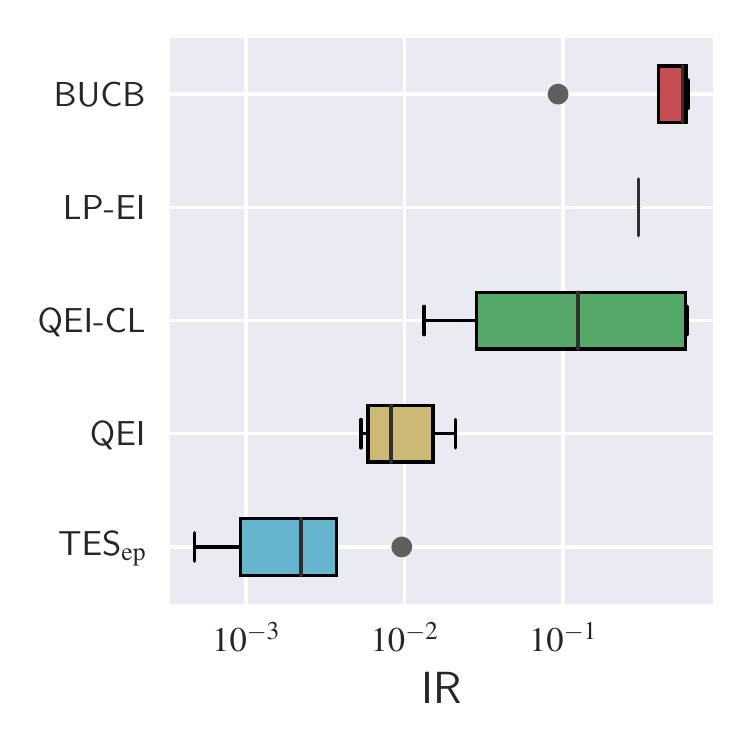}
	\\
	(g) log10P $|\mcl{B}| = 10$.
	&
	(h) log10P $|\mcl{B}| = 20$.
	\\
	\multicolumn{2}{c}{
	\includegraphics[height=0.2\textwidth]{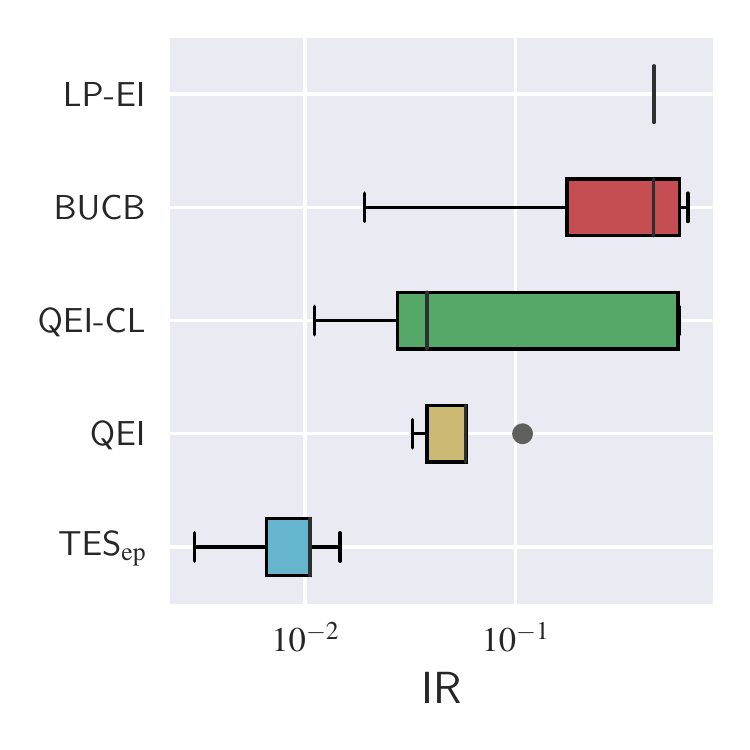}
	}
	\\
	\multicolumn{2}{c}{
	(i) log10P $|\mcl{B}| = 40$.
	}
\end{tabular}
\caption{Box plots of batch BO (i.e., $|\mcl{B}| > 1$) results for the GP sampled function, Hartmann-4D, and log10P.}
\label{fig:batchboxplot}
\end{figure}

Additionally, the box plots of IR in the last iteration of random experimental runs are shown in Figs.~\ref{fig:expsyncboxplot},~\ref{fig:realworldboxplot},~and~\ref{fig:batchboxplot}.

\end{document}